\documentclass{article}

\PassOptionsToPackage{numbers, compress}{natbib}
\usepackage{arxiv_style}

\usepackage[utf8]{inputenc} %
\usepackage[T1]{fontenc}    %
\usepackage{hyperref}       %
\usepackage{url}            %
\usepackage{booktabs}       %
\usepackage{amsfonts}       %
\usepackage{nicefrac}       %
\usepackage{microtype}      %
\usepackage[table]{xcolor}

\title{Early Failure Detection and Intervention in\\Video Diffusion Models}

\author{
{Byungki Kwon}$^{1}$ \quad
{Sohwi Lim}$^{2}$ \quad
{Nam Hyeon-Woo}$^{1}$ \quad
{Moon Ye-Bin}$^{1}$ \quad
{Tae-Hyun Oh}$^{2}$ \\
\\
$^{1}$Pohang University of Science and Technology (POSTECH) \\
$^{2}$Korea Advanced Institute of Science and Technology (KAIST)
}

\usepackage{comment}
\usepackage{enumitem}
\usepackage{multirow}
\usepackage{wrapfig}
\usepackage{colortbl}
\usepackage[normalem]{ulem}

\usepackage{caption}
\usepackage{graphicx}

\setlist[itemize]{align=parleft,left=0pt}

\definecolor{kwon}{rgb}{0.36, 0.54, 0.66}
\definecolor{kjs}{rgb}{0.784, 0.003, 0.313}
\definecolor{azure(colorwheel)}{rgb}{0.0, 0.5, 1.0}
\definecolor{nicegreen}{rgb}{0.0, 0.7, 0.1}
\definecolor{ashblue}{rgb}{0.36, 0.54, 0.66}
\definecolor{ashgrey}{rgb}{0.7, 0.75, 0.71}
\definecolor{applegreen}{rgb}{0.55, 0.71, 0.0}
\definecolor{jy}{rgb}{0.58, 0, 0.827}
\definecolor{cornellred}{rgb}{0.7, 0.11, 0.11}
\definecolor{darkcyan}{rgb}{0.0, 0.55, 0.55}
\definecolor{CuGray}{gray}{0.9}
\definecolor{airforceblue}{rgb}{0.36, 0.54, 0.66}
\definecolor{rev}{rgb}{0.784, 0.003, 0.313}
\definecolor{pink}{cmyk}{0, 0.7808, 0.4429, 0.1412}
\definecolor{amethyst}{rgb}{0.6, 0.4, 0.8}
\definecolor{black}{rgb}{0.0, 0.0, 0.0}
\definecolor{tb3_yellow}{rgb}{0.996, 1.0, 0.6}
\definecolor{tb3_orange}{rgb}{0.980, 0.8, 0.604}
\definecolor{tb3_red}{rgb}{0.972, 0.6, 0.6}
\definecolor{dimgray}{rgb}{0.41, 0.41, 0.41}
\definecolor{brickred}{rgb}{0.8, 0.25, 0.33}
\definecolor{bleudefrance}{rgb}{0.19, 0.55, 0.91}
\definecolor{blue(ncs)}{rgb}{0.265, 0.445, 0.765}
\definecolor{blue(ryb)}{rgb}{0.01, 0.28, 1.0}
\definecolor{cyan}{rgb}{0.0, 1.0, 1.0}
\definecolor{darkbrown}{rgb}{0.4, 0.26, 0.13}
\definecolor{brown(traditional)}{rgb}{0.59, 0.29, 0.0}
\definecolor{hyos}{rgb}{0.662, 0.482, 0.960}
\definecolor{magenta}{rgb}{0.98, 0.176, 0.815}

\newcolumntype{g}{>{\columncolor{CuGray}}c}
\newcolumntype{z}{>{\columncolor{CuGray}}l}

\renewcommand{\paragraph}[1]{\vspace{1mm}\noindent\textbf{#1.}\,\,}

\usepackage{xspace}

\makeatletter
\def\@fnsymbol#1{\ensuremath{\ifcase#1\or *\or \dagger\or \ddagger\or
   \mathsection\or \mathparagraph\or \|\or **\or \dagger\dagger
   \or \ddagger\ddagger \else\@ctrerr\fi}}
\makeatother

\def\onedot{.\@\xspace}
\def\eg{\emph{e.g}\onedot}

\newcommand{\be}{\begin{eqnarray}}
\newcommand{\ee}{\end{eqnarray}}
\newcommand{\bee}{\begin{eqnarray*}}
\newcommand{\eee}{\end{eqnarray*}}

\newcommand{\matrixb}{\left[ \begin{array}}
\newcommand{\matrixe}{\end{array} \right]}

\newcommand{\empha}[1]{{\color{red} \bf #1}}

\definecolor{green(ncs)}{rgb}{0.0, 0.62, 0.42}

\usepackage{amssymb}
\usepackage{pifont}
\usepackage{makecell}

\usepackage{amsmath}
\usepackage{orcidlink}
\usepackage[ruled,vlined]{algorithm2e}
\DontPrintSemicolon
\usepackage{algpseudocode}
\usepackage[most]{tcolorbox}
\usepackage{listings}

\lstdefinestyle{promptstyle}{
    basicstyle=\ttfamily\scriptsize,
    breaklines=true,
    breakatwhitespace=false,
    columns=fullflexible,
    keepspaces=true,
    showstringspaces=false,
    frame=none
}

\let\cite\citep

\begin{document}

\maketitle

\begin{center}
  \centering
\captionsetup{type=figure}
\includegraphics[width=\linewidth]{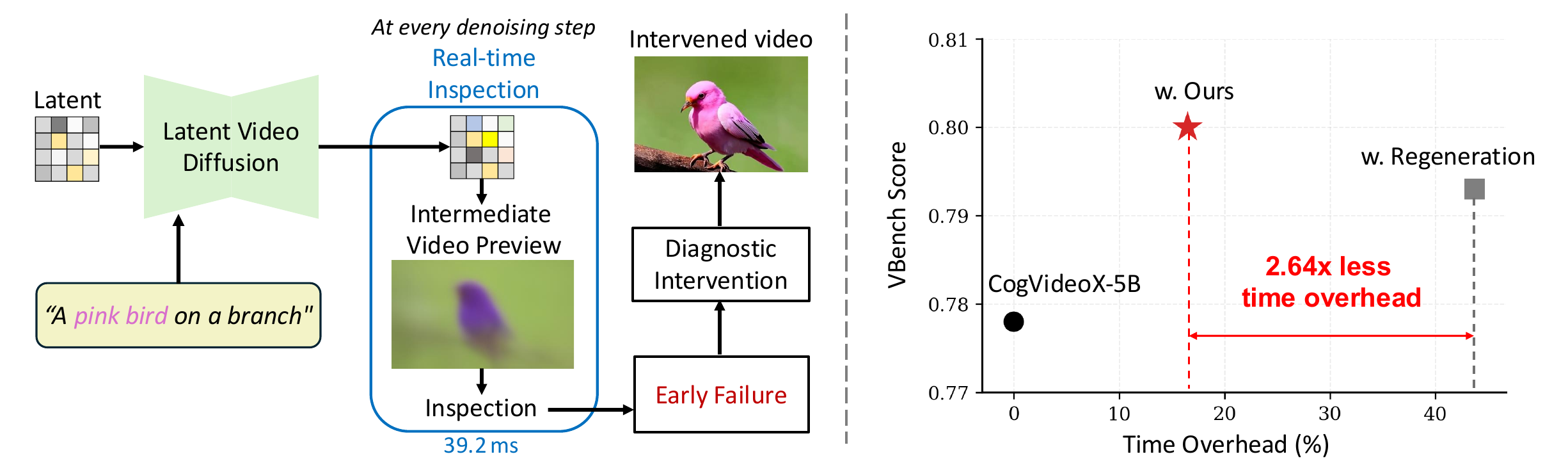}
  \captionof{figure}{\textbf{Real-time inspection enables early failure detection in latent video diffusion.} By decoding intermediate latents at every denoising step, our method identifies failing trajectories early and selectively intervenes. Compared to post-hoc regeneration, we achieve comparable VBench scores with 2.64× lower time overhead.}
  \label{fig:teaser}
\end{center}

\begin{abstract}
Text-to-video (T2V) diffusion models have rapidly advanced, yet generations still occasionally fail, forcing users into trial-and-error regeneration that wastes substantial compute and energy. 
While early failure detection has been explored for image diffusion, it remains unaddressed in on-the-fly, single-trajectory video diffusion
due to the prohibitive cost of decoding spatiotemporal latents at every step. The VAE decoder of CogVideoX-5B takes approximately 4\,s per call on an A100, comparable to a denoising step, rendering per-step inspection impractical.
We propose an \textit{early failure detection and intervention pipeline} for video diffusion that makes real-time, step-wise RGB inspection of sampling tractable. Our approach rests on a key empirical observation: \textit{the final video quality is largely predictable within the first $\sim$10 of 50 denoising steps}, and is reliably detectable from intermediate latents once they are decoded into RGB and scored by standard vision--language evaluators.
To exploit this on the fly, we introduce a \textit{Real-time Inspection} (RI) module that scores intermediate previews in 39.2\,ms, over 100$\times$ faster than a CogVideoX-5B denoising step. 
The key enabler is a \textit{Latent-to-RGB} (L2R) converter (0.059\,M parameters, 19.7\,ms) showing that learned spatial upsampling alone without channel expansion suffices to produce plausible video previews 200$\times$ faster than the native VAE while preserving the semantic structure needed for scoring.
When a failure is predicted, a diagnostic intervention framework hierarchically applies preview-driven corrections, escalating compute only when needed.
Experiments on CogVideoX-5B and Wan2.1-1.3B show consistent improvements and a favorable cost-quality trade-off against full-video feedback regeneration, a strong post-hoc baseline that inspects complete videos, with $\sim$2.5$\times$ lower additional time overhead.
Our method is also orthogonal to prompt refinement and training-free sampling guidance.
Project page: \url{https://byungki-k.github.io/EarlyFailure/}

\end{abstract}

\section{Introduction}\label{sec:intro}
Text-to-video (T2V) diffusion models~\cite{yang2024cogvideox, wan2025wan, kong2024hunyuanvideo, chen2025goku, blattmann2023stable, liu2024sora, hacohen2024ltx, bar2024lumiere, wiedemer2025video} have rapidly advanced, yet generations still occasionally fail with weak text--video alignment, poor perceptual quality, or temporal inconsistency~\cite{jan2025text, huang2024vbench, liu2023discovering, xing2024survey, sun2025t2v}. Such failures arise from missing semantic coverage or unlucky sampling trajectories~\cite{shi2025imbalance, rodriguez2025coral, samuel2024generating, xu2025good, ramesh2025test}, and are difficult to predict in advance, 
incurring substantial compute and energy cost as users iterate toward a satisfactory output~\cite{ma2025scaling}.

A natural alternative is to monitor sampling and intervene early. For image diffusion, this has been explored via efficient inspection of intermediate latents in latent or model-internal space, either to evaluate alignment~\cite{xie2025dymo, ramos2025beyond, wang2025diffusion} or to steer generation~\cite{ahn2024self, ma2025scaling, jain2025diffusion, ramesh2025test}. The same approach is not directly viable for video: the native VAE decoder of CogVideoX-5B takes approximately 4\,s per intermediate latent on an A100, comparable to a single denoising step. Prior video-domain work either defers inspection until after full generation~\cite{choi2025we} or distributes it across parallel trajectories~\citep{liu2025video, oshima2025inference, baraldi2025verifier}, leaving on-the-fly, single-trajectory failure detection largely impractical.

We address this gap by enabling on-the-fly failure detection during video diffusion sampling. Our approach is grounded in a key empirical observation: when intermediate latents are decoded into RGB and scored by an off-the-shelf vision--language evaluator, the resulting score reliably predicts final video quality within the first $\sim$10 of 50 denoising steps. While early structural convergence has been visualized in image diffusion as a generation property~\cite{liu2024faster, yi2024towards}, prior on-the-fly intervention systems rely on model-internal signals such as cross-attention maps~\cite{chefer2023attend, cui2026diffusion}, denoiser activations~\cite{guo2026toward}, or small latent-space evaluators~\cite{papalampidi2025dynamic}; even systems that decode predicted latents to pixel space combine them with cross-attention features and a custom-trained hallucination predictor rather than off-the-shelf VLM scoring~\cite{betti2024optimizing}. Concurrent video-domain work targets adjacent but distinct goals: inference-time scaling that ranks parallel candidates with fine-tuned verifiers~\cite{baraldi2025verifier}, and human-in-the-loop interactive previews via lightweight multi-branch preview decoders~\cite{hong2025diffusionbrowser}, neither addressing on-the-fly, automated, single-trajectory failure detection within a per-step latency budget. We provide the first system that operates in this regime, and design it to make zero-shot, off-the-shelf VLM scoring on RGB previews actionable within that budget.

To realize this, we propose an \emph{early failure detection and intervention pipeline} with two components: a \emph{Real-time Inspection (RI)} module that scores intermediate previews in 39.2\,ms ($100\times$ faster than a denoising step of CogVideoX-5B), enabled by a \emph{Latent-to-RGB (L2R) converter} (0.059\,M parameters, 19.7\,ms; $200\times$ faster than the native VAE); and a \emph{diagnostic intervention framework} that applies mode-specific corrections only when needed.

We validate on CogVideoX-5B~\cite{yang2024cogvideox} and Wan2.1-1.3B~\cite{wan2025wan}, achieving consistent VBench gains at $\sim$2.5$\times$ lower time overhead than full-video feedback regeneration, a strong post-hoc baseline. The framework is orthogonal to prompt rewriting and sampling guidance, and remains effective on Wan2.1-14B at 720p, 81 frames. 
By substituting on-the-fly correction for post-hoc regeneration, our framework points toward reducing the compute and energy spent on unsatisfactory generations.
Our contributions: 
(i) the first \emph{single-trajectory, on-the-fly} failure detection and intervention system for video diffusion, operating within a per-step budget via zero-shot, off-the-shelf VLM scoring on RGB previews;
(ii) a 0.059\,M-parameter L2R converter showing that learned spatial upsampling without channel expansion suffices to produce plausible previews 200$\times$ faster than the native VAE;
and (iii) a favorable cost--quality trade-off against feedback regeneration, with orthogonal gains atop prompt rewriting and sampling guidance.

\section{Related work}

\paragraph{Evaluation and failure diagnosis in diffusion models}
Prior work analyzes intermediate diffusion states for image generation through model-internal signals such as cross-attention maps~\cite{chefer2023attend,cui2026diffusion}, denoiser activations~\cite{guo2026toward}, or small latent-space evaluators~\cite{papalampidi2025dynamic}, used either to evaluate alignment with the prompt~\cite{ramos2025beyond,xie2025dymo} or to steer sampling toward better outcomes~\cite{ahn2024self,ma2025scaling,jain2025diffusion,ramesh2025test}. HEaD~\cite{betti2024optimizing} combines intermediate cross-attention maps with decoded predicted-final-image previews, scored by a custom-trained hallucination predictor, to support early stopping; DiffDoctor~\cite{wang2025diffdoctor} trains artifact detectors for offline failure diagnosis. These approaches operate primarily in image latents or model-internal space, and their direct extension to video is hindered by the prohibitive cost of decoding spatiotemporal latents at every step (Sec.~\ref{sec:intro}).
In video diffusion, prior work treats failure handling as either a post-hoc problem~\cite{choi2025we} or a multi-candidate search problem that maintains parallel trajectories with intermediate evaluation~\cite{liu2025video,oshima2025inference}. Extending this search paradigm, \citet{baraldi2025verifier} characterize the temporal coherence of VLM-based verifiers on decoded intermediate videos and propose verifier fine-tuning for inference-time scaling, while \citet{hong2025diffusionbrowser} take a different direction with lightweight multi-branch preview decoders for human-interactive guidance. Concurrent works further pursue the search paradigm: LatSearch~\cite{zhao2026latsearch} learns a latent reward model for reward-guided resampling, and \citet{tang2026seeking} score parallel candidates at intermediate checkpoints. In contrast, we operate on a single trajectory with zero-shot, off-the-shelf VLM scoring, requiring no trained reward model or verifier, enabling on-the-fly intervention with 20.3\% additional overhead.

\paragraph{Sampling guidance for video diffusion}
Following classifier-free guidance~\citep{ho2022classifier}, several methods steer video diffusion sampling by exploiting temporal characteristics: modifying attention~\citep{ahn2024self}, skipping spatiotemporal layers~\citep{hyung2025spatiotemporal}, suppressing noise variance~\citep{luo2025enhance}, or guiding flow coherence~\citep{shaulov2025flowmo}. While effective on average, these methods do not guarantee success on every prompt or sampling trajectory. Our pipeline is orthogonal to them: it can be combined with any of these guidance methods to selectively intervene only on the residual failure cases.

\begin{figure}[t]
  \centering
  \includegraphics[width=0.93\linewidth]{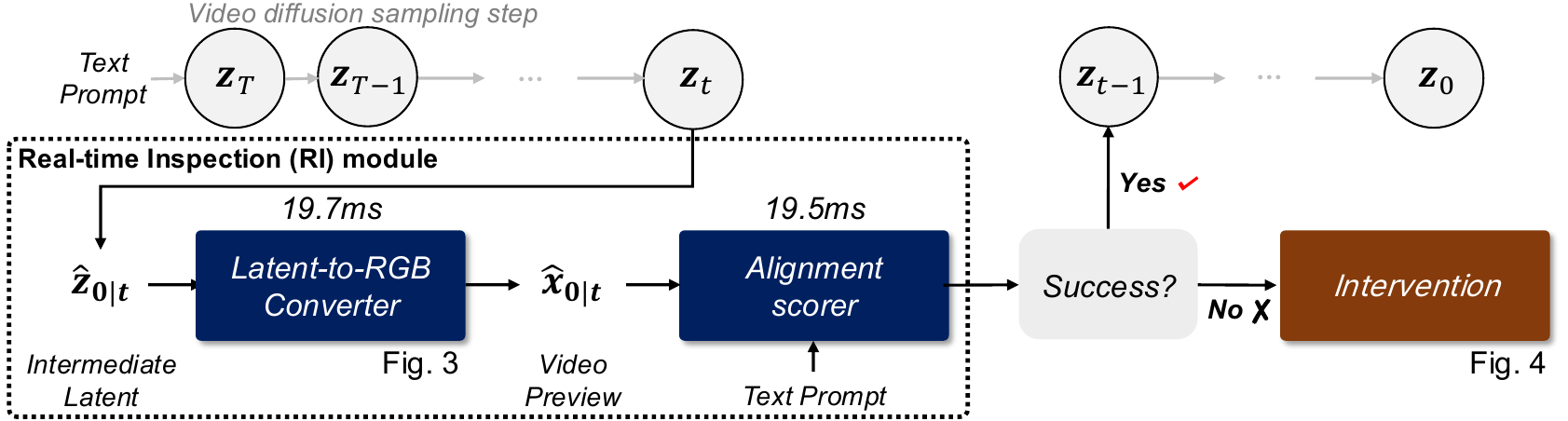}
    \caption{\textbf{Early failure detection and intervention pipeline.} 
    The Real-time Inspection (RI) module converts predicted clean latents into RGB video previews and evaluates their semantic alignment score with the input prompt (19.7\,ms + 19.5\,ms). Intervention is triggered based on the score.
    }
\label{fig:converter_arch}
\vspace{-3mm}
\end{figure}

\section{Method}
We address failure handling in latent text-to-video diffusion as a system-level problem under a strict per-step latency budget. The native VAE decoder of CogVideoX-5B, for instance, costs approximately 4\,s per call on an A100, comparable to a single denoising step, so any inspection mechanism that operates inside the sampling loop must be substantially cheaper than this. Our framework (Fig.~\ref{fig:converter_arch}) is built around two components designed to fit this budget: a Real-time Inspection (RI) module (Sec.~\ref{sec:early_detection}) that scores intermediate previews in 39.2\,ms, and a diagnostic intervention framework (Sec.~\ref{sec:intervention}) that escalates compute only when a failure is predicted. Together they replace full-video feedback regeneration with cheap per-step inspection (39.2\,ms per step) and selective intervention on flagged samples, keeping the dataset-averaged overhead within 20.3\%.

\subsection{Real-time Inspection (RI) module}
\label{sec:early_detection}
The RI module operationalizes the empirical premise discussed above (validated in Sec.~\ref{sec:exp-why}, Fig.~\ref{fig:early-pred}): RGB-decoded previews enable plug-in use of pretrained VLMs in two roles, an off-the-shelf scorer for failure detection and a reasoner for Visual Feedback Intervention (Sec.~\ref{sec:intervention}); neither role is available in latent space without a learned reward model~\cite{zhao2026latsearch}. We realize this within the per-step budget through three components introduced below: a Latent-to-RGB (L2R) converter that produces previews $200\times$ faster than the native VAE, an alignment scorer (19.5\,ms), and a dynamic failure detector that adapts to per-sample semantic convergence.

\paragraph{Latent-to-RGB (L2R) converter}
We design L2R as a purpose-built converter for the inspection role, rather than reusing any existing decoder. Its design rests on two principles from prior work: latent-to-RGB mapping can be largely linear~\citep{metzer2023latent}, and learned upsampling restores spatial detail more faithfully than bilinear interpolation~\citep{teed2020raft}.

\begin{figure}[t]
  \centering
  \includegraphics[width=\linewidth]{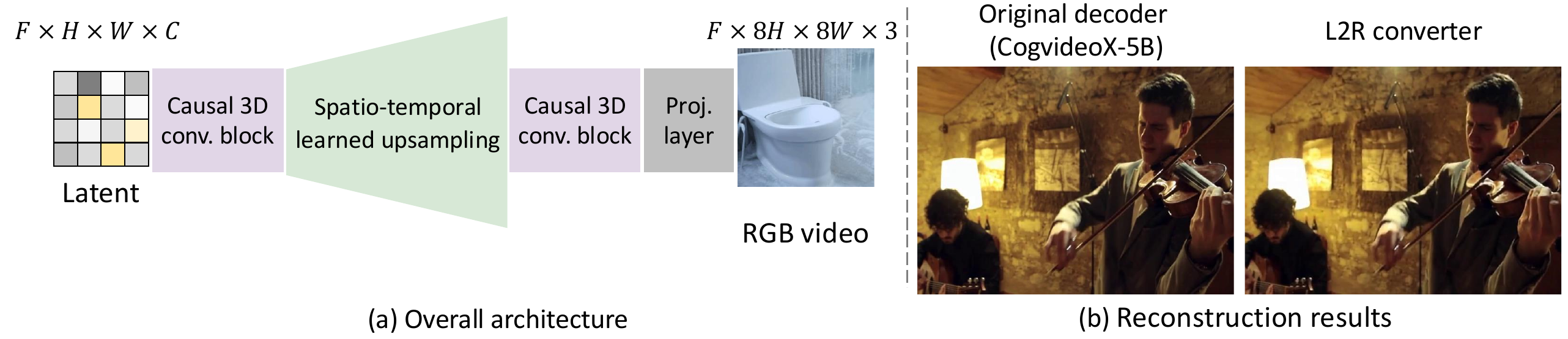}
    \caption{\textbf{Real-time Latent-to-RGB (L2R) converter.} 
    (a) Our L2R architecture maps latents to an RGB video using causal 3D convolutional blocks, spatio-temporal learned upsampling, and a projection layer. 
    It runs in 19.7\,ms with only 0.059\,M parameters.
    (b) The reconstruction result shows
    reasonable visual quality compared to the result produced by the original decoder.}
  \label{fig:arch_detail}
\end{figure}

Building on these principles, our architecture (Fig.~\ref{fig:arch_detail}) makes two deliberate omissions justified by the inspection role. First, \textbf{no channel expansion}: unlike standard latent video decoders that expand intermediate features several-fold to recover RGB fidelity, all our intermediate features stay at the latent channel dimension and reduce to three channels only at the final projection layer. Second, \textbf{no temporal upsampling}: video quality metrics are more sensitive to spatial than temporal resolutions (Appendix~\ref{sec:C.5}), so we leave the temporal axis at the latent's frame count. The remaining components are minimal: two causal 3D convolution blocks for temporal context and a spatio-temporal learned upsampling module that recovers the $8\times$ spatial compression, followed by the projection layer.
The result is a 0.059\,M-parameter module that produces video previews for 480p, 49-frame video generations in 19.7\,ms, over $200\times$ faster than the native CogVideoX VAE, with $1700\times$ smaller parameter memory (0.138\,MB vs.\ 242.5\,MB; Table~\ref{tab:supple_recon_quality}).
L2R achieves reconstruction quality comparable to or better than TinyAE~\cite{BoerBohan2025TAEHV} while approaching Turbo-VAED~\cite{zou2025turbo}, using $190\times$ and $690\times$ fewer parameters respectively (Appendix~\ref{sec:A.2}).
This confirms our design hypothesis: for preview-based inspection rather than pixel-perfect reconstruction, learned spatial upsampling alone, without channel expansion, suffices to produce plausible previews while preserving the semantic structure needed for VLM-based scoring (validated in Sec.~\ref{sec:exp-why}). We provide the training details in Appendix~\ref{sec:A.2}.

\paragraph{Alignment scorer}
Within the per-step latency budget, the choice of scorer is otherwise flexible.
We adopt ViCLIP~\cite{wang2023internvid} as our default for two reasons:
(i) unlike image--text models, ViCLIP is purpose-built for video--text alignment
and captures temporal semantics relevant to video previews, and
(ii) it is sufficiently lightweight to fit our per-step budget.
Our framework is not tied to ViCLIP: Sec.~\ref{sec:exp-robust} validates the pipeline
with OpenCLIP, indicating that the scorer is
a swappable component rather than a hard dependency.
To meet the budget, we move frame preprocessing and similarity computation onto
the GPU and cache the text embedding once per sample, since the prompt is fixed
throughout sampling. Together these optimizations reduce scoring to 19.5\,ms per call, comparable to L2R itself. We denote by $s_k$ the alignment score it produces at denoising step $k$.

\paragraph{Dynamic failure detector}
Diffusion sampling is known to converge sample-dependently~\citep{huang2024vbench, ma2024deepcache}, and we observe the same effect in semantic content (Appendix~\ref{sec:A.3}): some samples crystallize their content within a few steps while others remain ambiguous longer. A fixed-step predictor would either miss late-converging samples or waste budget on early-converging ones. We instead aggregate the score sequence $\{s_k\}$ over a short inspection window and emit a single predicted final score $\tilde{s}_0$ via a sample-adaptive policy based on the score's level, recent trend, and local stability, so that confident trajectories exit early and ambiguous ones continue until their scores stabilize (full rules in Appendix~\ref{sec:A.3}).
A sample is classified as a failure if $\tilde{s}_0 < \tau$. We set $\tau = 0.22$ for ViCLIP based on the score distribution of \emph{real} videos (Fig.~\ref{fig:threshold}), and validate this choice with a 26-participant user study at score levels 0.18, 0.22, and 0.26. The same procedure yields $\tau = 0.30$ when OpenCLIP is used as the scorer (Appendix~\ref{sec:A.5}).
Across both models, the detector reaches the dynamic decision at step 11/50 on average for CogVideoX and 10/50 for Wan2.1.

\begin{figure}[t]
  \centering
  \includegraphics[width=.95\linewidth]{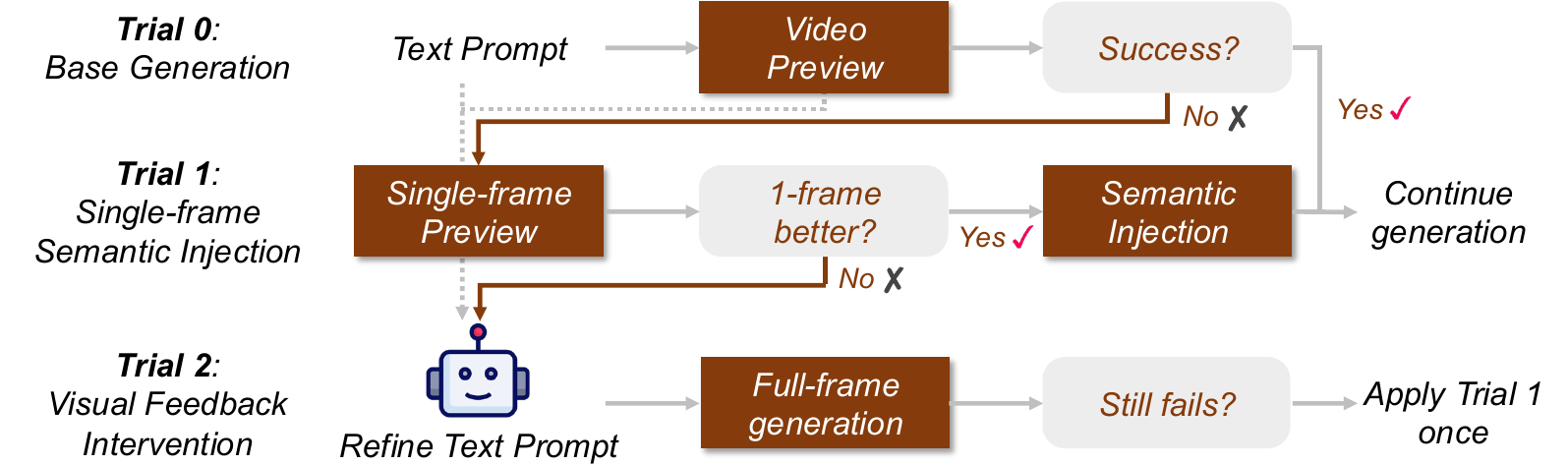}
  \caption{\textbf{Diagnostic intervention framework.} If the generation is predicted to succeed, Trial 0 continues to generate video without intervention. When failure is predicted, the next Trials are applied, escalating from inexpensive to more costly fixes. Trial 1 performs a single-frame probe and injects its latent as a semantic anchor; Trial 2 invokes a VLM as a visual feedback reasoner over the failed preview to refine the prompt. If Trial 2 also fails, Trial 1 is attempted once more.}
  
    \label{fig:intervene}
\vspace{-3mm}
\end{figure}

\subsection{Diagnostic intervention framework}
\label{sec:intervention}

When the detector flags a sample, we apply a hierarchical sequence of interventions, escalating compute only when needed (Fig.~\ref{fig:intervene}). The order, from cheapest to most expensive, is: continue the base generation (Trial~0); inject a single-frame semantic anchor (Trial~1); refine the prompt from the failed preview with a VLM and restart (Visual Feedback Intervention, Trial~2); and apply Trial~1 on top of the refined prompt if it still fails (Trial~2$\rightarrow$1). Branching is observation-driven: each transition is decided from the preview signals already available at the inspection step, so most samples terminate before reaching the more expensive trials. Even in the worst case, the cumulative cost is about 56\% of a full second generation (Fig.~\ref{fig:cost_stair}).

\noindent\textbf{Trial 0. Base Generation.}
If the detector predicts $\tilde{s}_0 \geq \tau$, generation continues with no modification.

\noindent\textbf{Trial\,1.\,\,Single-frame Semantic Injection.}
Trial~1 targets a specific failure mode: the model captures what the prompt describes in a single frame, but fails to extend it into a coherent video.
We diagnose this from the previews themselves. Using the same T2V model with the temporal dimension collapsed, we generate a single-frame preview of the prompt and score its predicted alignment $\tilde{s}_{\mathrm{img}}$ (the single-frame analog of $\tilde{s}_0$) with the same dynamic detector. When $\tilde{s}_{\mathrm{img}} > \tilde{s}_0 + \delta$ (with
$\delta = 0.05$), the single frame is better aligned than the video,
supporting the temporal-failure hypothesis. The margin $\delta$ guards
against borderline cases where the single-frame and video scores differ
only by detector noise, ensuring that Trial~1 is invoked only when
single-frame generation is substantively stronger.
We then reuse the single-frame intermediate prediction $\hat{\mathbf{z}}^{\mathrm{img}}_{0 \mid  k_{\mathrm{img}}}$ as a semantic anchor: it is broadcast along the temporal axis and blended into the video latent over the first one or two denoising steps using a soft mask matched to each step's noise level (Algorithm~\ref{alg:single_frame_injection}). Because single-frame generation is orders of magnitude cheaper than
video, the entire trial completes in seconds, well below the cost of
Trial~2. We therefore try Trial~1 first: if the single-frame test
indicates a temporal failure, the injection corrects the sample;
otherwise, we move on to Trial~2.

\noindent\textbf{Trial\,2.\,\,Visual Feedback Intervention.}
When the single-frame test fails ($\tilde{s}_{\mathrm{img}} \leq \tilde{s}_0 + \delta$), the failure is unlikely to be purely temporal, and Trial~1 is bypassed. Instead, we directly leverage the second affordance unlocked by RGB-space inspection: a pretrained VLM as a reasoner over the failed preview itself. The VLM receives the original prompt and the decoded preview, identifies the visible mismatch, and returns an intent-preserving refined prompt with an optional negative prompt. We restart generation under the refined prompt and monitor it with the same RI module. If the new run still satisfies $\tilde{s}_0 < \tau$, the failure likely combines semantic and temporal causes, and we apply Trial~1 once on top (Trial~2$\rightarrow$1) before terminating.

The VLM prompt (Fig.~\ref{fig:prompt_template_refinement}) and per-Trial examples (Appendix~\ref{sec:D}) are provided in the Appendix.

\section{Experiments}
\label{sec:experiments}
We evaluate the proposed early failure detection and
intervention pipeline on widely used open-source video diffusion
models, CogVideoX-5B and Wan2.1-1.3B. 
We further evaluate Wan2.1-14B at 720p with 81 frames in Section~\ref{sec:exp-robust}.
Section~\ref{sec:exp-setup} describes
the setup. Section~\ref{sec:exp-main} reports the main cost--quality
trade-off against full-video feedback regeneration. 
Section~\ref{sec:exp-robust} shows the gains hold under an alternative internal scorer and an additional evaluation benchmark.
Section~\ref{sec:exp-orth} confirms orthogonality to prompt refinement
and training-free sampling guidance. Section~\ref{sec:exp-why}
empirically grounds the design choices behind the RI module, and
analyzes what the detector and
intervention do.

\subsection{Experimental setup}
\label{sec:exp-setup}

\noindent\textbf{Models and inference.}
We use CogVideoX-5B~\citep{yang2024cogvideox} and
Wan2.1-1.3B~\citep{wan2025wan} with their official codebases, default
seeds, and default guidance scales. Unless otherwise stated, all
generations use 50 denoising steps to produce 480p, 49-frame videos.
For paired comparison, generations flagged as early failures are also
run to full completion without any intervention, isolating the effect
of our intervention from run-to-run stochasticity caused by GPU
non-determinism.

\noindent\textbf{Benchmarks and metrics.}
Unless noted otherwise, all experiments use VBench~\citep{huang2024vbench}.
Crucially, the text--video alignment metric in VBench (\textit{Overall Consistency}) is computed against the \emph{original} prompts, regardless of any VLM-based prompt refinement applied during generation, ensuring that our prompt refinement does not artificially inflate alignment scores.
We additionally evaluate with VideoScore2~\citep{he2025videoscore2}, 
a learned video evaluator trained independently of VBench (Sec.~\ref{sec:exp-robust}).

\noindent\textbf{VLM and scorer.}
We use GPT-5-mini as the default VLM for visual feedback
intervention and ViCLIP as the default alignment scorer
in the RI module. We additionally evaluate Qwen3.8-27B
and OpenCLIP as replacements for the VLM and scorer,
respectively (Table~\ref{tab:vbench_regen}).

\begin{table}[t]
\centering
\caption{\textbf{Comparison with regeneration across two prompt
regimes.} We compare against full-video feedback regeneration
using original user prompts (\emph{Original prompt}) and
LLM-rewritten prompts (\emph{Prompt rewriting}).
Ours achieves competitive or better VBench score
with $\sim2.5\times$ less additional time overhead.
Shaded rows denote component variants, replacing the default
VLM (GPT-5-mini) with Qwen3.8-27B or the alignment scorer
(ViCLIP) with OpenCLIP.
Bold indicates the better result between Regeneration and
default Ours, excluding variants.}
\label{tab:vbench_regen}

\resizebox{\linewidth}{!}{
\begin{tabular}{l cccc cccc}
\toprule
& \multicolumn{4}{c}{\textbf{CogVideoX-5B}}
& \multicolumn{4}{c}{\textbf{Wan2.1-1.3B}} \\
\cmidrule(lr){2-5} \cmidrule(lr){6-9}
& \multicolumn{3}{c}{\textbf{VBench Score}}
& \textbf{Time}
& \multicolumn{3}{c}{\textbf{VBench Score}}
& \textbf{Time} \\
\cmidrule(lr){2-4} \cmidrule(lr){6-8}
\textbf{Method}
& \textbf{Final}
& \textbf{Quality}
& \textbf{Semantic}
& \textbf{Overhead (\%)}
& \textbf{Final}
& \textbf{Quality}
& \textbf{Semantic}
& \textbf{Overhead (\%)} \\
\midrule

Original prompt
& 0.778 & 0.803 & 0.680 & 0.00
& 0.789 & 0.811 & 0.700 & 0.00 \\

+ Regeneration
& 0.793 & 0.813 & 0.716 & 43.73
& \textbf{0.815} & \textbf{0.823} & \textbf{0.783} & 51.25 \\

+ Ours
& \textbf{0.800} & \textbf{0.816} & \textbf{0.735} & \textbf{16.55}
& 0.805 & 0.817 & 0.756 & \textbf{20.30} \\

\rowcolor{black!8}
\quad\emph{w/ Qwen3.8-27B (VLM)}
& 0.791 & 0.812 & 0.706 & 16.49
& 0.802 & 0.820 & 0.732 & 20.08 \\

\rowcolor{black!8}
\quad\emph{w/ OpenCLIP (scorer)}
& 0.799 & 0.813 & 0.741 & 15.75
& 0.810 & 0.826 & 0.749 & 20.44 \\

\midrule

Prompt rewriting
& 0.787 & 0.818 & 0.663 & 3.93
& 0.815 & 0.820 & 0.796 & 11.70 \\

+ Regeneration
& 0.801 & 0.821 & 0.722 & 48.87
& \textbf{0.822} & \textbf{0.824} & \textbf{0.815} & 53.74 \\

+ Ours
& \textbf{0.813} & \textbf{0.822} & \textbf{0.766} & \textbf{18.70}
& 0.820 & 0.822 & 0.809 & \textbf{25.12} \\

\bottomrule
\end{tabular}
}
\end{table}

\subsection{Main results: cost--quality trade-off}
\label{sec:exp-main}

\noindent\textbf{Comparison with full-video feedback regeneration.}
We compare against full-video feedback regeneration, a strong post-hoc baseline that, upon detecting a failure, observes the complete video, uses a VLM to refine the prompt, and regenerates from scratch. This baseline has strictly more visual information than our method, which intervenes based on early previews. 

We evaluate at two operating points (Table~\ref{tab:vbench_regen}): raw user prompts (\textit{Original prompt}) and LLM-rewritten prompts (\textit{Prompt rewriting}). Our pipeline establishes a Pareto-favorable operating point on the cost--quality frontier across both models and both prompt regimes, with $2.5\times$ lower time overhead than Regeneration. On CogVideoX-5B, this cheaper operating point further surpasses Regeneration on Final in both regimes (0.800 vs.\ 0.793; 0.813 vs.\ 0.801). On Wan2.1-1.3B, Ours closely matches Regeneration under Prompt rewriting (0.820 vs.\ 0.822); under raw prompts, Regeneration's $+0.010$ Final advantage requires $2.5\times$ more time overhead, leaving Ours strictly Pareto-optimal at the cheap end of the frontier. Because our system intervenes on only ${\sim}40\%$ of samples by design, the dataset-averaged Final gains ($+0.022$ on CogVideoX-5B, $+0.016$ on Wan2.1-1.3B) understate the per-sample effect: on the intervened subset, the per-sample VBench Final gain reaches ${\sim}{+}0.054$ on CogVideoX-5B, all without model tuning. By construction, such recovery cannot be achieved by methods acting uniformly across all samples.

\begin{figure}[t]
  \centering
  \includegraphics[width=\linewidth]{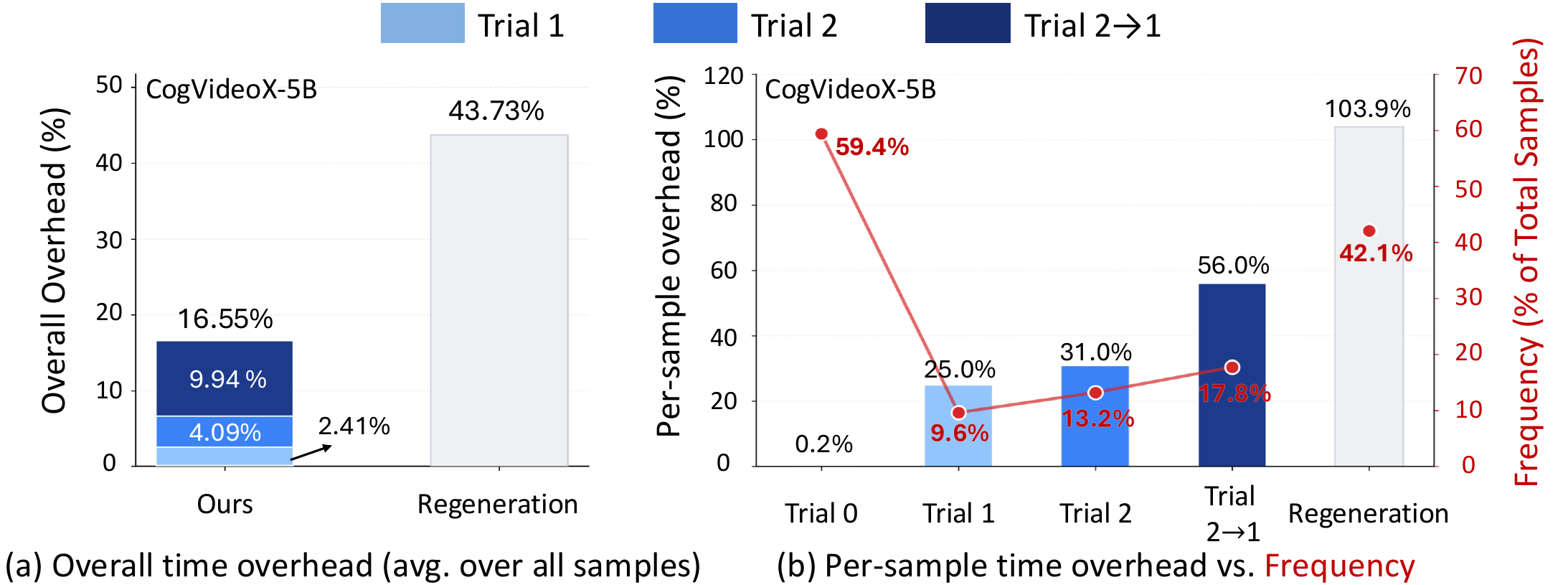}
\caption{\textbf{Efficiency of selective early intervention.} (a) Overall time overhead relative to base generation. Ours incurs only +16.6\% additional overhead, substantially lower than the regeneration of failure samples (+43.7\%). (b) Per-sample cost increases with deeper trials, yet remains cheaper than regeneration. Since deeper trials are triggered for only a subset of samples (red curve), the overall time overhead stays modest.}
  \label{fig:cost_stair}
\vspace{-3mm}
\end{figure}

\paragraph{Where the savings come from}
Figure~\ref{fig:cost_stair} explains the source of efficiency. Per-sample 
overhead grows with intervention depth (Fig.~\ref{fig:cost_stair}b), but only 
a small fraction of samples reach the more expensive trials, while most either 
complete without intervention or are corrected early. As a result, the 
dataset-averaged overhead remains below $20.3\%$ on both models 
(Fig.~\ref{fig:cost_stair}a). Including VLM cost, regeneration incurs a 
per-sample overhead of $103.9\%$ relative to base generation. The Wan2.1-1.3B 
breakdown is provided in Appendix~\ref{sec:B.1}.

\begin{table}[t]
\centering
\small
\caption{\textbf{Evaluation on VideoScore2.} Ours
consistently improves all three dimensions over the baseline, with
substantially larger gains on the predicted-failure subset.
VideoScore2 scores \emph{Visual Quality},
\emph{Text--Video Alignment}, and
\emph{Physical Consistency} on a 1--5 scale.}
\label{tab:videoscore2}
\setlength{\tabcolsep}{5pt}
\resizebox{0.95\linewidth}{!}{
\begin{tabular}{lcccccc}
\toprule
& \multicolumn{3}{c}{CogVideoX-5B} & \multicolumn{3}{c}{Wan2.1-1.3B} \\
\cmidrule(lr){2-4} \cmidrule(lr){5-7} 
Setting & Visual & T--V Align. & Phys. & Visual & T--V Align. & Phys. \\
\midrule
Original prompt (baseline)        & 3.86 & 3.82 & 3.66 & 3.70 & 3.73 & 3.54 \\
Original prompt + Ours            & \textbf{3.93}\,(+0.07) & \textbf{3.96}\,(+0.14) & \textbf{3.74}\,(+0.08)
                                   & \textbf{3.82}\,(+0.12) & \textbf{3.86}\,(+0.13) & \textbf{3.62}\,(+0.08) \\
\midrule
Predicted Failure (no intervene)       & 3.77 & 3.55 & 3.56 & 3.59 & 3.54 & 3.49 \\
Predicted Failure + Ours          & \textbf{3.94}\,(+0.17) & \textbf{3.91}\,(+0.35) & \textbf{3.74}\,(+0.18)
                                   & \textbf{3.84}\,(+0.25) & \textbf{3.82}\,(+0.27) & \textbf{3.67}\,(+0.18) \\
\bottomrule
\end{tabular}}
\end{table}
\begin{table}[t]
\centering
\caption{\textbf{Orthogonality to existing sampling enhancements.}
Our method consistently improves VBench scores across both
CogVideoX-5B and Wan2.1-1.3B when applied on top of two training-free
sampling enhancements: Enhance-A-Video~\cite{luo2025enhance} and
Spatiotemporal Skip Guidance (STG)~\cite{hyung2025spatiotemporal}.
The gains come from early detection and selective intervention,
which operate orthogonally to these enhancement techniques.}
\label{tab:vbench_full}
\resizebox{0.85\linewidth}{!}{
\begin{tabular}{@{} l ccc ccc @{}}
\toprule
\multirow{2}[2]{*}{\textbf{Method}} & \multicolumn{3}{c}{\textbf{CogVideoX-5B}} & \multicolumn{3}{c}{\textbf{Wan2.1-1.3B}} \\
\cmidrule(lr){2-4} \cmidrule(lr){5-7}
& \textbf{Final} & \textbf{Quality} & \textbf{Semantic} & \textbf{Final} & \textbf{Quality} & \textbf{Semantic} \\
\midrule
Original prompt (baseline) & 0.778 & 0.803 & 0.680 & 0.789 & 0.811 & 0.700 \\
\quad + Ours & \textbf{0.800} & \textbf{0.816} & \textbf{0.735} & \textbf{0.805} & \textbf{0.817} & \textbf{0.756} \\
\midrule
Original prompt + Enhance-A-Video & 0.766 & 0.794 & 0.656 & 0.791 & 0.812 & 0.705 \\
\quad + Ours & \textbf{0.790} & \textbf{0.804} & \textbf{0.736} & \textbf{0.805} & \textbf{0.817} & \textbf{0.756} \\
\midrule
Original prompt + STG & 0.784 & 0.811 & 0.679 & 0.794 & 0.822 & 0.683 \\
\quad + Ours & \textbf{0.802} & \textbf{0.822} & \textbf{0.722} & \textbf{0.806} & \textbf{0.825} & \textbf{0.728} \\
\bottomrule
\end{tabular}
}
\end{table}

\begin{figure}[t]
  \centering
  \includegraphics[width=0.95\linewidth]{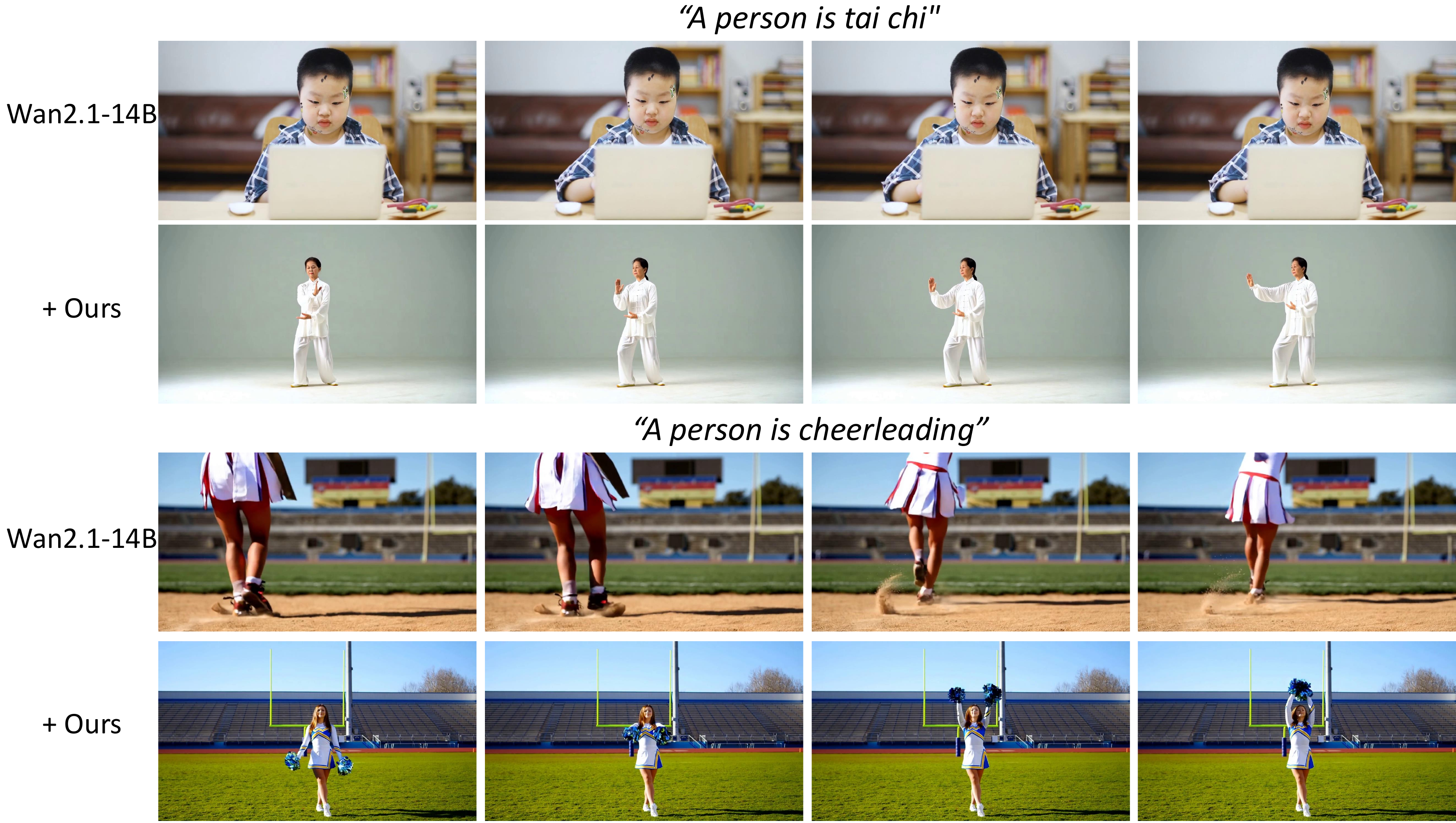}
\caption{\textbf{Qualitative results in a larger-model regime.}
On Wan2.1-14B at 720p with 81 frames, our method detects failures at an early stage and intervenes before full generation, avoiding the computation that would otherwise be wasted on completing failed trajectories.}

\label{fig:wan14b_qual}
\vspace{-3mm}
\end{figure}

\begin{table}[t]
\centering
\caption{\textbf{VBench evaluation on Wan2.1-14B at 720p,
81 frames.} We evaluate on the full set of 946 VBench prompts.
Our method improves semantic alignment with only 8.17\%
estimated additional time overhead.
Bold indicates the better VBench score.}
\label{tab:wan14b_vbench}

\begin{tabular}{l cccc}
\toprule
& \multicolumn{3}{c}{\textbf{VBench Score}}
& \textbf{Time} \\
\cmidrule(lr){2-4}
\textbf{Method}
& \textbf{Final}
& \textbf{Quality}
& \textbf{Semantic}
& \textbf{Overhead (\%)} \\
\midrule
Original prompt (baseline)
& 0.812 & 0.823 & 0.767 & 0.00 \\
+ Ours
& \textbf{0.819}
& \textbf{0.825}
& \textbf{0.795}
& 8.17 \\
\bottomrule
\end{tabular}
\end{table}

\subsection{Generality of the gains}
\label{sec:exp-robust}
We examine the generality of our framework across evaluation metrics,
component choices, and a more demanding setting with a larger model,
higher resolution, and longer video sequences.

\noindent\textbf{Additional evaluation with VideoScore2.} We further evaluate with VideoScore2~\citep{he2025videoscore2}, a learned video evaluator that, given a generated video and its text prompt, returns three human-aligned scores on a 1--5 scale: \emph{Visual Quality} (perceptual fidelity), \emph{Text--Video Alignment} (prompt faithfulness), and \emph{Physical Consistency} (plausibility of motion and interactions). Importantly, VideoScore2 is trained independently of VBench, providing an external check on our gains. Table~\ref{tab:videoscore2} reports results on the full VBench evaluation set. Our method consistently improves all three dimensions over the baseline on both models. The gains are substantially larger on the predicted-failure subset (\eg, $+0.35$ in Text--Video Alignment for CogVideoX-5B), confirming that selective intervention is particularly effective on failure-prone samples.

\noindent\textbf{Modularity of the alignment scorer.} The scorer is a swappable component of the RI module rather than a hard dependency. We adopt ViCLIP as the default for two reasons: it is purpose-built for video--text alignment, and it is lightweight enough to fit the per-step latency budget. To demonstrate this modularity, we further evaluate the pipeline with OpenCLIP, an image--text model unrelated to VBench's evaluation and not designed for video. The \emph{Ours~(Scorer: OpenCLIP)} row in Table~\ref{tab:vbench_regen} shows comparable VBench gains at similar overhead on both models, indicating that future, more capable alignment scorers can be plugged in as drop-in replacements without re-engineering the pipeline.

\noindent\textbf{Extension to a larger-model regime.}
We further evaluate our framework on Wan2.1-14B at 720p with
81 frames, using the full set of 946 VBench prompts.
Since Wan2.1-1.3B and Wan2.1-14B share the same VAE,
we reuse the same L2R converter trained on the Wan2.1
VAE latent space without retraining.
We also retain the default ViCLIP scorer and failure threshold
$\tau=0.22$.
As shown in Table~\ref{tab:wan14b_vbench}, our method improves
the VBench Final score from 0.812 to 0.819, primarily driven
by improved semantic alignment (0.767 to 0.795), while
maintaining comparable visual quality (0.823 to 0.825).
At 720p, each denoising step takes 67.05 s for
81-frame video generation and 1.29~s for single-frame
generation. Our method incurs only 8.17\% estimated
additional time overhead because of the increased cost of
full-video denoising.
Figure~\ref{fig:wan14b_qual} shows qualitative examples where our method detects failures early and intervenes before full generation, reducing the resources spent on unsuccessful trajectories.

\begin{figure}[t]
  \centering
  \includegraphics[width=\linewidth]{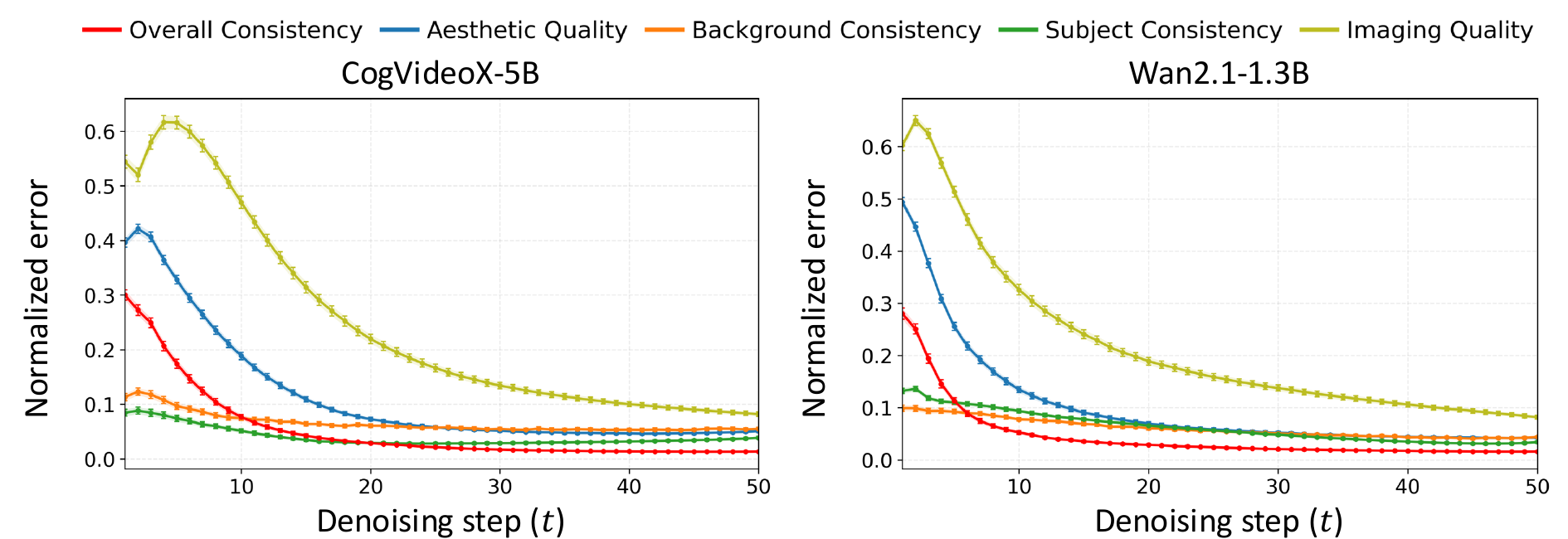}
\caption{\textbf{Early predictability of final video quality.}
For 1{,}800 prompts on CogVideoX and Wan2.1, we report MAE between VBench metrics computed on intermediate L2R previews and on final videos decoded by the original VAE. Errors drop sharply by $\sim$10 steps, notably for Overall Consistency.}
  \label{fig:early-pred}
\vspace{-3mm}
\end{figure}

\subsection{Orthogonality to existing enhancements}
\label{sec:exp-orth}

We examine whether our intervention remains beneficial when combined
with existing enhancement techniques. Orthogonality with prompt
rewriting is already evidenced in Table~\ref{tab:vbench_regen}: applying our
pipeline on top of LLM-rewritten prompts further improves Final
score beyond the prompt-rewriting baseline alone (0.787$\to$0.813
on CogVideoX-5B; 0.815$\to$0.820 on Wan2.1-1.3B). Here we further
integrate our pipeline with two recent training-free sampling
enhancements, Enhance-A-Video~\citep{luo2025enhance} and
Spatiotemporal Skip Guidance (STG)~\citep{hyung2025spatiotemporal}.
Table~\ref{tab:vbench_full} summarizes the results. Across both
models and both enhancements, our method consistently improves
all scores, confirming that early detection and
intervention operate orthogonally to existing sampling guidance
techniques.

\subsection{Empirical analysis of early predictability}
\label{sec:exp-why}

We quantitatively analyze the conditions that make on-the-fly failure detection feasible. Three findings jointly ground our pipeline: (i) final video quality is largely predictable from RGB-decoded intermediate latents within the first $\sim$10 of 50 denoising steps, enabling intervention on incomplete trajectories; (ii) a single alignment signal, \emph{Overall Consistency}, captures quality variations across other VBench metrics, keeping the per-step inspection budget tight; and (iii) the failure threshold $\tau = 0.22$ aligns with both the real-video score distribution and human perception.

\noindent\textbf{Early predictability of final video quality.}
A core premise of our pipeline is that final video quality can be reliably estimated from early denoising steps. We validate this on 1{,}800 prompts (946 from VBench plus 854 from Panda70M~\citep{chen2024panda}) across both models, computing at every denoising step the \emph{normalized error}, defined as MAE divided by each metric's dynamic range, between L2R previews and final VAE-decoded videos. Fig.~\ref{fig:early-pred} shows that prediction error plateaus within the first $\sim$10 steps, indicating that intermediate scores are reasonable proxies for final quality.

\noindent\textbf{Overall Consistency as a proxy for video quality.}
For real-time inspection, we use Overall Consistency as the sole alignment signal to fit the 19.5\,ms budget. A balanced binary classification on top/bottom quantiles shows PR-AUC above 0.5 across all VBench metrics, confirming that this single signal captures diverse failure modes (Appendix~\ref{sec:C.6}).

\begin{figure}[t]
  \centering
  \includegraphics[width=\linewidth]{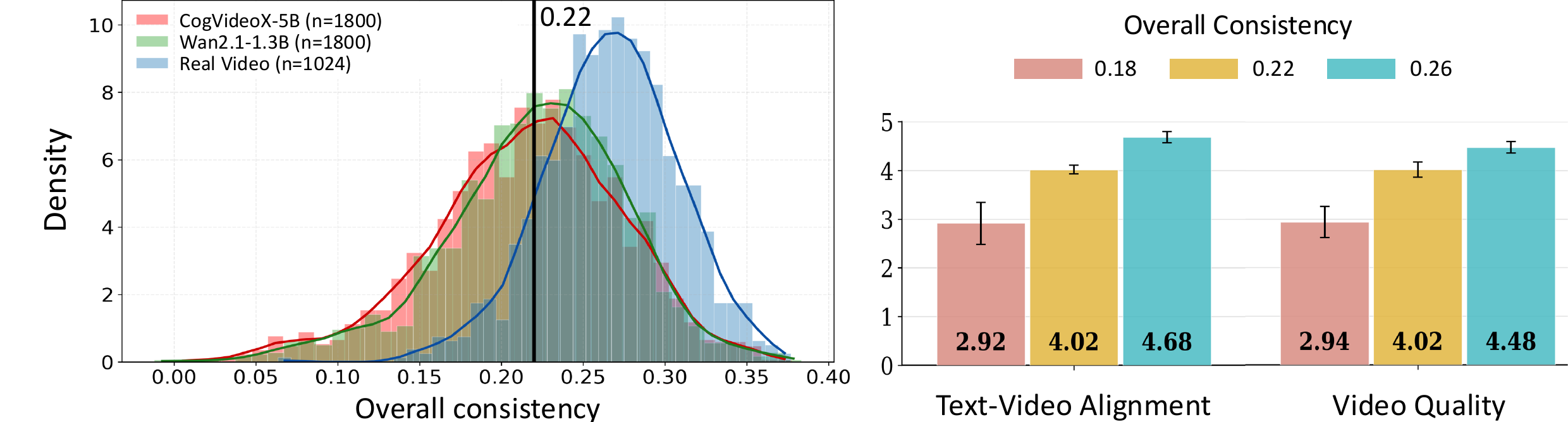}
\caption{\textbf{Selecting the failure threshold $\tau$ for Overall Consistency.}
\textbf{Left:} Distributions of Overall Consistency for generated and real videos. Real videos concentrate at higher scores; we set $\tau=0.22$ so that the retained generated samples better match the real-video regime.
\textbf{Right:} A human study with 26 participants over 30 samples (780 responses) validates the perceptual boundary: scores of 0.18 are often judged as failures, whereas scores$\ge$0.22 are considered reasonable.
}
\label{fig:threshold}
\vspace{-3mm}
\end{figure}

\noindent\textbf{Threshold selection for failure detection.}
We set $\tau$ by combining distributional evidence with a human study (Fig.~\ref{fig:threshold}). Real videos (1{,}024 Pexels clips, captioned by GPT-5-mini and scored with the same alignment scorer~\cite{wang2023internvid}) concentrate at higher \textit{Overall Consistency} than T2V outputs, which exhibit a pronounced low-score tail; $\tau = 0.22$ removes this tail. The framework remains effective across a range of $\tau$ (Appendix~\ref{sec:C.8}), so $\tau$ acts as a practical quality--efficiency knob.

\begin{figure}[t]
  \centering
  \includegraphics[width=\linewidth]{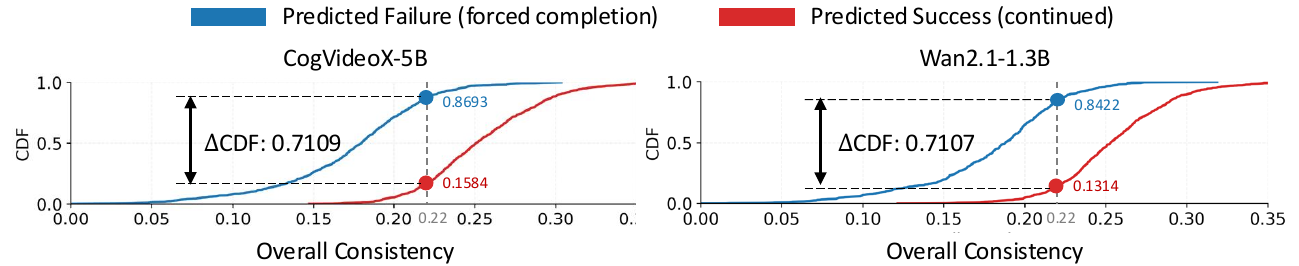}
    \caption{\textbf{Accuracy of early failure detection.}
    Final Overall Consistency distributions for samples predicted to pass or fail (failures forced to complete without intervention). Strong separation across models demonstrates reliable early discrimination.}
  \label{fig:detector}
\vspace{-3mm}
\end{figure}

\noindent\textbf{Behavior of detection and intervention.}
The dynamic failure detector reaches a decision at step 11/50 (CogVideoX-5B) and 10/50 (Wan2.1-1.3B) on average. Running flagged samples to completion \emph{without intervention} yields cleanly separated final \textit{Overall Consistency} ($\Delta\mathrm{CDF}\!\approx\!0.71$; Fig.~\ref{fig:detector}), confirming early discrimination.

\section{Conclusion}
We present the first early failure detection and intervention pipeline for latent text-to-video diffusion that operates on a single trajectory within a strict per-step latency budget. The key enabler is a Real-time Inspection (RI) module which decodes intermediate latents into RGB via a 0.059\,M-parameter Latent-to-RGB converter and scores them with an off-the-shelf, zero-shot vision--language evaluator in 39.2\,ms, over $100\times$ faster than a single denoising step.
This makes step-wise RGB inspection practical for on-the-fly, single-trajectory video diffusion sampling, and corroborates our key empirical finding that final video quality is largely predictable from the first $\sim$10 of 50 denoising steps.
When a failure is predicted, a diagnostic intervention framework escalates computation hierarchically, replacing post-hoc regeneration with on-the-fly correction at approximately $2.5\times$ lower additional time overhead under original prompts.
Across CogVideoX-5B and Wan2.1-1.3B at 480p with 49 frames, as well as Wan2.1-14B at 720p with 81 frames, the framework delivers consistent quality gains while remaining orthogonal to prompt refinement and sampling guidance.
Building on these results, detecting finer-grained failures at earlier stages, together with more efficient and effective intervention strategies, represents a promising direction toward reducing wasted generation compute for service providers and shortening the time to satisfactory outputs for users.

\paragraph{Broader impact} Text-to-video generation often requires multiple rounds of regeneration before users obtain a satisfactory result, with each failed attempt consuming substantial GPU compute and energy. By identifying and correcting failures within a single trajectory rather than running them to completion or restarting from scratch, our method directly reduces the compute and energy spent on unsatisfactory generations. As an efficiency contribution orthogonal to generative capability, it does not introduce new misuse vectors beyond those already present in the underlying T2V models.

\paragraph{Limitations and future work} While our framework supports interchangeable lightweight alignment scorers, validated with both ViCLIP (default) and OpenCLIP, subtle failures such as fine-grained motion artifacts or compositional errors fall below the discriminative resolution of current lightweight evaluators and remain a target for future work. More expressive yet efficient video evaluators would extend the failure modes our framework can detect and recover from.

\begin{ack}
This work was supported by the National Research Foundation of Korea (NRF) funded by the Korean government (MSIT) in 2026 (Grant No. RS-2024-00451947, Project Title: Interaction Framework for Connecting Users to Generative Experiences via Time-Space Expanded Multimedia: Centering on Social Communication Disorders (45\%)), the Institute of Information \& Communications Technology Planning \& Evaluation (IITP) grant funded by the Korean government (MSIT) (No. RS-2026-25518317, Development of AI memory mechanism that reflects human cognitive principles (40\%); No. 2022-0-00124, No. RS-2022-II220124, Development of Artificial Intelligence Technology for Self-Improving Competency-Aware Learning Capabilities (15\%)), the InnoCORE program of the Ministry of Science and ICT (26-InnoCORE-01; N10250156), the Artificial Intelligence Industrial Convergence Cluster Development Project funded by the Ministry of Science and ICT (MSIT, Korea) \& Gwangju Metropolitan City, and the Institute of Information \& Communications Technology Planning \& Evaluation (IITP) grant funded by the Korean government (MSIT) (No. RS-2019-II191906, Artificial Intelligence Graduate School Program (POSTECH)).

\end{ack}

\bibliographystyle{plainnat}
\bibliography{main}

\clearpage

\appendix

\section*{Technical appendices and supplementary material}
We provide implementation details of our early failure detection 
and intervention pipeline, and describe how the time overhead is computed in detail.
In addition, we present a range of additional experiments, including the multi-seed evaluation, 
overhead comparison with test-time scaling methods, comparison with a Best-of-3 
baseline, threshold sensitivity analysis, and disaggregated analysis of each 
intervention branch. 

{\hypersetup{hidelinks}
\section*{Contents}

\noindent\hyperref[sec:A]{\textbf{A \ \ \ Implementation details}}\\
\hyperref[sec:A.1]{\text{\qquad A.1 \ \ \ Experiment details}} \\ 
\hyperref[sec:A.2]{\text{\qquad A.2 \ \ \ Real-time Inspection (RI) module}} \\ 
\hyperref[sec:A.3]{\text{\qquad A.3 \ \ \ Dynamic failure detector}} \\
\hyperref[sec:A.4]{\text{\qquad A.4 \ \ \ Diagnostic intervention framework}} \\
\hyperref[sec:A.5]{\text{\qquad A.5 \ \ \ Threshold calibration for OpenCLIP}} \\
\hyperref[sec:B]{\textbf{B \ \ \ Analysis of time overhead}} \\
\hyperref[sec:B.1]{\text{\qquad B.1 \ \ \ Efficiency of selective early intervention on Wan2.1-1.3B}}\\
\hyperref[sec:B.2]{\text{\qquad B.2 \ \ \ Computation of time overhead}}\\ 
\hyperref[sec:C]{\textbf{C \ \ \ Additional experiments}} \\
\hyperref[sec:C.2]{\text{\qquad C.1 \ \ \ Qualitative results of our pipeline}} \\ 
\hyperref[sec:C.3]{\text{\qquad C.2 \ \ \ Multi-seed evaluation on VBench}} \\ 
\hyperref[sec:C.4]{\text{\qquad C.3 \ \ \ Overhead comparison with test-time scaling methods}} \\ 
\hyperref[sec:C.5]{\text{\qquad C.4 \ \ \ Spatiotemporal sensitivity of VBench metrics}} \\ 
\hyperref[sec:C.6]{\text{\qquad C.5 \ \ \ Overall Consistency as a proxy for video quality}} \\ 
\hyperref[sec:C.7]{\text{\qquad C.6 \ \ \ Comparison with a best-of-3 baseline}} \\ 
\hyperref[sec:C.8]{\text{\qquad C.7 \ \ \ Threshold sensitivity of early failure detection}} \\ 
\hyperref[sec:C.9]{\text{\qquad C.8 \ \ \ Effectiveness of intervention on failure cases}} \\ 
\hyperref[sec:C.10]{\text{\qquad C.9 \ Disaggregated analysis of intervention effects}} \\ 
\hyperref[sec:D]{\textbf{D \ \ \ Additional qualitative results}} \\
\noindent\rule{\linewidth}{0.2pt}
}

\hypersetup{linkcolor=blue}

\section{Implementation details}\label{sec:A}

\subsection{Experiment details}\label{sec:A.1}
For VBench evaluation, we generate one video per text prompt using the same inference settings and seed across all methods. For CogVideoX-5B~\cite{yang2024cogvideox} and Wan2.1-1.3B~\cite{wan2025wan}, we use the default seed and guidance scale provided in the official codebases.
To enable comparison with base generation, we force the sampling process to continue full completion without any intervention, even when early failures are flagged. This allows direct comparison against the base generation under the same evaluation setting.
Following prior work~\cite{huang2024vbench}, the text--video alignment metric in VBench, \textit{Overall Consistency}, is computed using the original user prompts, regardless of any VLM-based prompt refinement applied during generation.
We also report VBench results across all dimensions in Tab.~\ref{tab:vbench_per_dim}. 

\paragraph{Discussion of per-dimension VBench results}
Table~\ref{tab:vbench_per_dim} reports our method's effect across all 16 VBench 
dimensions. The improvements are broad: our method improves the majority of 
dimensions on both models, with substantial gains on semantic dimensions such 
as Human Action ($+0.090$ on CogVideoX-5B; $+0.140$ on Wan2.1-1.3B), Multiple 
Objects ($+0.161$; $+0.067$), Object Class ($+0.051$; $+0.072$), and Scene 
($+0.079$; $+0.109$), confirming that selective early intervention effectively 
corrects alignment failures. Beyond alignment, perceptual and temporal 
dimensions also improve consistently, including Subject Consistency, 
Background Consistency, Motion Smoothness, and Imaging Quality, indicating 
that the gains generalize beyond the alignment-driven failure detection signal.
The one dimension that does not improve is Dynamic Degree, a binary 
classification of whether each video exceeds an optical-flow threshold 
estimated by RAFT~\cite{teed2020raft}; the fraction of videos crossing 
this threshold is preserved on CogVideoX-5B ($0.542 \rightarrow 0.542$) 
but decreases on Wan2.1-1.3B ($0.542 \rightarrow 0.403$).
Table~\ref{tab:dynamic_degree} decomposes this trade-off across four settings 
and shows that the decrease is largely inherited from the VLM-based prompt 
refinement step rather than introduced by our intervention itself: applying 
VLM-refined prompts alone already reduces this fraction substantially on 
Wan2.1-1.3B ($0.542 \rightarrow 0.319$), and full-video feedback regeneration, 
which uses the feedback-based refinement step, exhibits a comparable drop 
($0.542 \rightarrow 0.431$). Notably, on CogVideoX-5B our method preserves 
this fraction better than regeneration ($0.542$ vs.\ $0.458$) while requiring 2.64$\times$ lower time overhead.

\subsection{Real-time Inspection (RI) module}\label{sec:A.2}
The Real-time Inspection (RI) module is designed to monitor the denoising process of T2V diffusion models in real time. To do so, it first converts intermediate latent representations into RGB video previews, and then evaluates how well the current generation aligns with the input text using an alignment scorer. In this way, the RI module provides an online estimate of semantic alignment during sampling. In the following, we describe the implementation details of the Latent-to-RGB (L2R) converter and the alignment scorer used in the RI module.

\begin{figure}[t]
  \centering
  \includegraphics[width=0.95\linewidth]{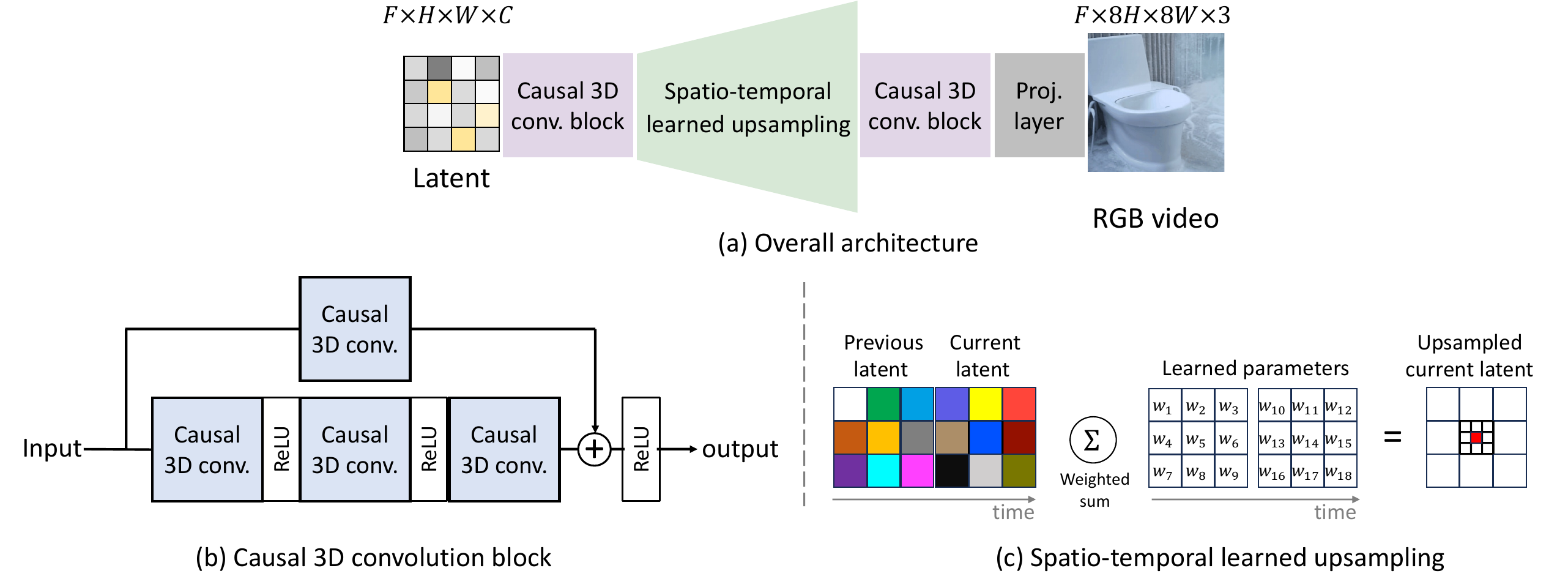}
  \caption{\textbf{Architecture of Latent-to-RGB (L2R) converter.}
(a) Overall architecture.
The converter takes a latent video tensor as input and produces an RGB video preview through two causal 3D convolution blocks, a spatio-temporal learned upsampling, and a projection layer.
(b) Causal 3D convolution block with a main branch and a skip branch.
(c) Illustration of spatio-temporal learned upsampling for a single \empha{red target} location.
Given the previous and current latents, learned parameters are used to predict the upsampled latent at the current step by weighted summation. In practice, this operation is implemented densely as a 3D convolution over the latent.}
  \label{fig:supple_l2r}
\end{figure}

\paragraph{Latent-to-RGB (L2R) converter}
The design of our Latent-to-RGB (L2R) converter is primarily motivated by the need for an efficient preview module that balances low latency with moderate visual quality. Our architecture is informed by the observation in Latent-NeRF~\cite{metzer2023latent} that linear approximations are often sufficient for decoding latent representations into plausible RGB colors, as well as the findings in RAFT~\cite{teed2020raft} regarding the superiority of learned upsampling over bilinear interpolation for recovering fine-grained spatial details. 

Figure~\ref{fig:supple_l2r}(a) shows the overall architecture of the L2R converter.
The model consists of two causal 3D convolution blocks, one spatio-temporal learned upsampling module, and a final projection layer.
To keep the model lightweight, all hidden features use the same channel dimension as the input latent, and the channel dimension is reduced to RGB only at the final layer. As shown in Fig.~\ref{fig:supple_l2r}(b), each causal 3D convolution block is built with a main branch and a skip branch, followed by nonlinear activations.
We use a kernel size of $2 \times 3 \times 3$ for time, height, and width dimensions, which allows the network to incorporate temporal context from previous frames.
Figure~\ref{fig:supple_l2r}(c) illustrates how the spatio-temporal learned upsampling works for a single target location.
For clarity, we show a $3\times$ upsampling example.
Given the previous and current latents, the module predicts the target latent at the current step by taking a weighted sum with learned parameters.
In practice, this local operation is applied densely to all output locations and implemented efficiently as a 3D convolution.
Finally, the projection layer converts the refined latent features into RGB using two 2D convolution layers with kernel sizes $9\times9$ and $3\times3$.
\begin{table}[t]
\centering
\caption{\textbf{Efficiency and quality of latent-to-RGB converter.}
We measure PSNR on DAVIS and Pexels. Then, we measure the decoding time/memory at 480$\times$720.
L2R offers moderate PSNR with the lowest latency and memory, providing previews suitable for inspection for both models.}
\label{tab:supple_recon_quality}
\resizebox{\linewidth}{!}{%
\begin{tabular}{lcccccccccc}
\toprule
& \multicolumn{5}{c}{\textbf{CogVideoX-5B}} & \multicolumn{5}{c}{\textbf{Wan2.1-1.3B}} \\
\cmidrule(lr){2-6} \cmidrule(lr){7-11}
& & \textbf{DAVIS} & \textbf{Pexels} & & & & \textbf{DAVIS} & \textbf{Pexels} & & \\ 
\textbf{Decoder} 
& \textbf{Param. (M)} & \textbf{PSNR} $\uparrow$ & \textbf{PSNR} $\uparrow$ & \textbf{Time (s)} & \textbf{Mem (MB)}
& \textbf{Param. (M)} & \textbf{PSNR} $\uparrow$ & \textbf{PSNR} $\uparrow$ & \textbf{Time (s)} & \textbf{Mem (MB)} \\
\midrule
Original   
& 123.4 & \textbf{32.80} & \textbf{35.30} & 3.990 & 242.5 
& 73.30  & \textbf{32.68} & \textbf{35.41} & 1.702 & 140.4 \\
TinyAE      
& 11.32  & 26.56 & 27.35 & 0.116 & 21.60 
& 11.32  & 30.50 & 33.46 & 0.116 & 21.60 \\
Turbo-VAED 
& 40.69  & 32.03 & 35.03 & 0.599 & 114.6 
& -      & -     & - & -      & - \\
Ours       
& \textbf{0.059}   & 30.00 & 32.50 & \textbf{0.0197} & \textbf{0.138} 
& \textbf{0.059}   & 28.98 & 31.40 & \textbf{0.0197} & \textbf{0.138} \\
\bottomrule
\end{tabular}%
}
\end{table}

Training requires only the pretrained VAE encoder and is independent of the diffusion model.
We collect 33k real-world video clips from Pexels~\cite{Pexels} with sufficient temporal dynamics (mean optical-flow > 0.5 via RAFT), encode them into latents using the VAE encoder, and train on latent–RGB pairs.
Training completes in 11 hours on a single NVIDIA A100 GPU. 

Table~\ref{tab:supple_recon_quality} compares our L2R converter against the original decoders and efficient VAE baselines, including Turbo-VAED~\cite{zou2025turbo} and TinyAE~\cite{BoerBohan2025TAEHV}.
We train separate converters for CogVideoX and Wan2.1 to match their respective VAE latent spaces, and evaluate reconstruction quality on the DAVIS-2017~\cite{pont20172017} test split with videos truncated to 49 frames and resized to 480$\times$720. 
We additionally construct a Pexels test set consisting of 500 videos and compare reconstruction performance as well.
For L2R, 
PSNR is computed against ground-truth RGB frames under the same temporal subsampling setting. While the original decoders achieve higher PSNR, our L2R attains moderate PSNR with drastically reduced decoding time and memory usage.
Compared to efficient VAE baselines, L2R uses substantially fewer parameters and achieves faster inference while maintaining moderate reconstruction quality across both models, making it well-suited for inspection alongside generation.

\paragraph{Alignment scorer}
Within the per-step latency budget, the choice of scorer is otherwise flexible. 
We adopt ViCLIP~\cite{wang2023internvid} as our default for two reasons: 
(i)~unlike image--text models, ViCLIP is purpose-built for video--text alignment 
and captures temporal semantics relevant to video previews, and (ii)~it is 
sufficiently lightweight to fit our per-step budget. Our framework is not tied 
to ViCLIP: Sec.~4.3 of the main paper validates the pipeline with OpenCLIP as 
an alternative scorer, indicating that the alignment scorer is a swappable 
component rather than a hard dependency.

A natural concern is whether using ViCLIP at inspection time inflates VBench 
scores by construction. We argue this is not the case for two reasons. 
First, ViCLIP is used in only 2 of VBench's 16 dimensions (Overall Consistency 
and Temporal Style), yet our method improves the majority of dimensions on 
both models, with all but one dimension on Wan2.1-1.3B (Table~\ref{tab:vbench_per_dim}) preserved or improved. Second, gains transfer 
to VideoScore2~\cite{he2025videoscore2} (Sec.~4.3 of the main paper), a learned 
video evaluator trained independently of VBench and ViCLIP, with consistent 
improvements across Visual Quality, Text--Video Alignment, and Physical 
Consistency on both models. Together, these results indicate that the gains 
reflect genuine quality improvements rather than artifacts of using ViCLIP as 
the inspection signal.

Built upon the VBench implementation, we apply several implementation-level 
optimizations to meet the per-step latency budget for online use during 
denoising. We move the frame preprocessing and similarity computation to the 
GPU, and cache the text feature once per sample since the prompt remains fixed 
throughout sampling. Together these optimizations reduce scoring to only 
19.5\,ms per call, comparable to L2R itself. Note that we use the official 
VBench codebase for the final VBench scores reported in our experiments.

\begin{figure}[t]
  \centering
  \includegraphics[width=0.95\linewidth]{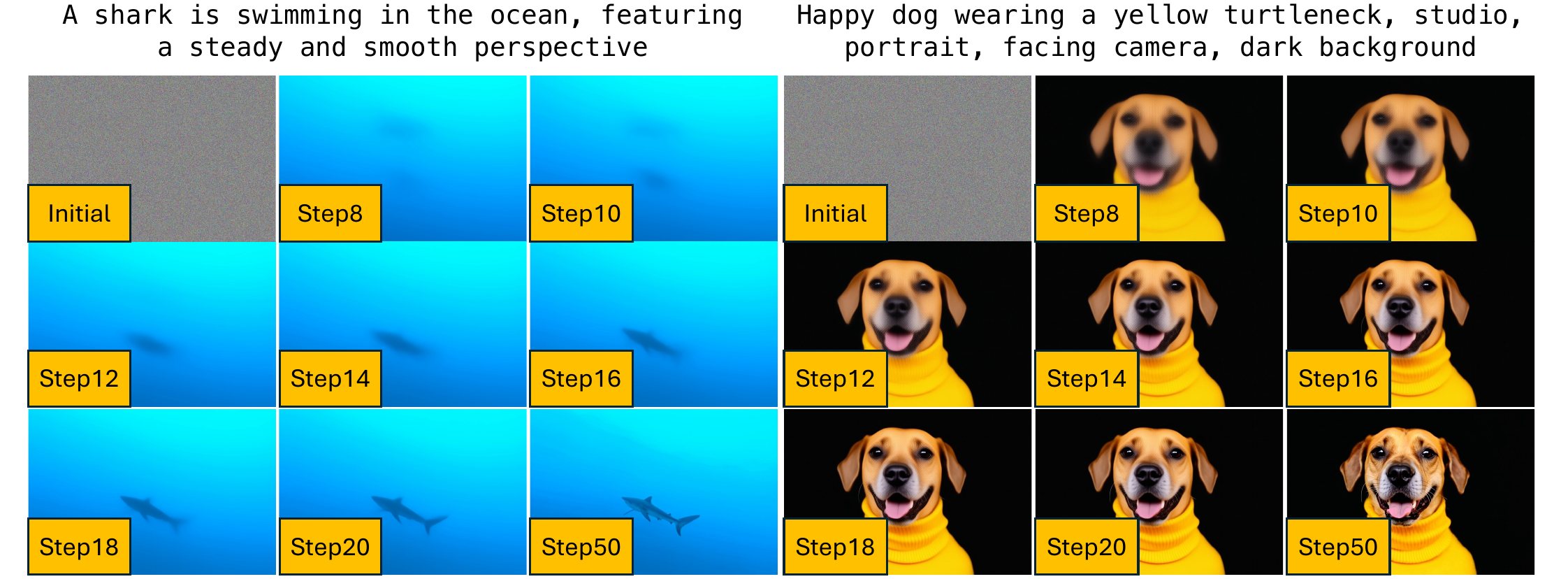}
\caption{\textbf{Semantic convergence varies across samples in video diffusion.}
We visualize the center frame (24th of 49) decoded from intermediate denoising steps for two text-to-video generations. While the dog sample exhibits recognizable semantics at an early stage, the shark sample remains ambiguous until later steps. This illustrates that semantic convergence in video diffusion is highly sample-dependent, motivating the dynamic failure detector rather than relying on a fixed early step.}
  \label{fig:supple_convergence}
\end{figure}

\subsection{Dynamic failure detector}\label{sec:A.3}
In Fig.~\ref{fig:early-pred} of the main paper, we show that several final video quality metrics are already predictable from early denoising steps.
Specifically, the error between the metric measured from an intermediate preview and that of the final video drops sharply within the first \(\sim 10\) denoising steps.

However, the convergence behavior of diffusion models is known to be sample-dependent~\cite{huang2024vbench,ma2024deepcache}. We further observe that semantic convergence in text-to-video generation also differs substantially across prompts, as shown in Fig.~\ref{fig:supple_convergence}. This suggests that monitoring only a fixed denoising step is less effective than using a sample-dependent monitoring strategy.

\begin{wraptable}{r}{0.5\linewidth}
\centering
\caption{\textbf{Overall Consistency between fixed-step vs.\ dynamic failure detector.}
Despite nearly identical average prediction steps, the dynamic strategy achieves lower normalized error on both models.
}
\resizebox{1.0\linewidth}{!}{   %
\begin{tabular}{@{} l l c c @{}}
\toprule
\textbf{Model} & \textbf{Method\quad} & \textbf{Avg. Step} & 
\makecell{\textbf{Normalized Error}\\\small ($\times 10^{-2}$ $\downarrow$)} \\
\midrule
\multirow{2}{*}{CogVideoX-5B\quad} 
& Fixed                  & 11 &  $6.67 \pm 0.45$ \\
& Dynamic      & 11.05 & $\mathbf{5.73 \pm 0.41}$ \\
\midrule
\multirow{2}{*}{Wan2.1-1.3B} 
& Fixed                 & 10 &  $5.24 \pm 0.34$ \\
& Dynamic       & 10.33 & $\mathbf{4.70 \pm 0.30}$ \\
\bottomrule
\end{tabular}
}
\label{tab:supple_dynamic_fixed}
\end{wraptable}

Since our RI module inspects a video preview in only 39.2\,ms (L2R + alignment scorer), we use a dynamic failure detector that monitors the alignment score at every denoising step.
Rather than making a decision from a single fixed step, the detector maintains a short history of intermediate scores and their step-wise changes, while also tracking the best score observed within a predefined inspection budget.

Our dynamic failure detector tracks the alignment trajectory and issues a sample-adaptive stopping decision based on its level, trend, and stability.
Let $\{s_k\}$ denote the alignment scores measured during denoising.
For the video preview inspection, the detector begins making decisions from step 8 and inspects the trajectory up to step 14.
Within this inspection window, it applies simple stopping rules based on the score level, recent trend, and local stability.
Specifically, it triggers \textit{early reject} when the score is low and remains nearly unchanged for several steps, \textit{give up} when the score is low and continues to decrease, \textit{early accept} when the score is already high and stable, and \textit{stable stop} when the score has sufficiently converged.
If none of these conditions are met within the inspection budget, the detector stops at the budget limit and returns the best score observed so far. We use the returned score as the predicted final alignment score, denoted by $\tilde{s}_0$.
A sample is then classified as successful if $\tilde{s}_0 \ge \tau$, and as a failure otherwise.

Table~\ref{tab:supple_dynamic_fixed} compares the dynamic failure detector with a fixed-step predictor under the same experimental setting as Fig.~\ref{fig:early-pred} in the main paper.
For each model, we use the average prediction step of the dynamic detector as the corresponding fixed step.
Even under this matched comparison, the dynamic detector achieves lower normalized error on CogVideoX-5B and Wan2.1-1.3B.
This result suggests that a sample-dependent stopping strategy provides a more accurate estimate of the final alignment score than monitoring a single fixed denoising step.
Notably, this improvement is achieved even though the detector includes failure-oriented policies such as \textit{early reject}, rather than focusing purely on final score prediction.
For single-frame preview inspection, we use a later inspection window, starting from step 20 and continuing up to step 30, since single-frame denoising progresses substantially faster than video denoising.

\begin{algorithm}[t]
\caption{Trial 1: Single-frame Semantic Injection}
\label{alg:single_frame_injection}
\begin{algorithmic}[1]
\Require Single-frame preview latent $\mathbf{z}^{\mathrm{img}}$, initial video latent $\mathbf{z}_T$, denoising timesteps $\{t_k\}_{k=2}^N$, text condition $c$, scheduler, T2V model, injection length $K=2$ \\
Expand $\mathbf{z}^{\mathrm{img}}$ with a temporal dimension
\State Set the reference noise to the initial latent: $\boldsymbol{\epsilon} \leftarrow \mathbf{z}_T$
\State Initialize the latent $\mathbf{z}$ at $t_2$: $\mathbf{z} \leftarrow \mathrm{AddNoise}(\mathbf{z}^{\mathrm{img}}, \boldsymbol{\epsilon}, t_2)$
\State \textbf{for} $k = 2, \dots, N$ \textbf{do}
\State \hspace{1em}\textbf{if} $k \leq K+1$ \textbf{then}
\State \hspace{2em}$\hat{\mathbf{z}} \leftarrow \mathrm{AddNoise}(\mathbf{z}^{\mathrm{img}}, \boldsymbol{\epsilon}, t_k)$
\State \hspace{2em}$w_k \leftarrow \mathrm{SoftMask}(t_k)$
\State \hspace{2em}$\mathbf{z} \leftarrow w_k \odot \hat{\mathbf{z}} + (1-w_k)\odot \mathbf{z}$
\State \hspace{1em}\textbf{end if}
\State \hspace{1em} $\mathbf{z}$ $\leftarrow$ Scheduler(T2V($\mathbf{z}, c, t_k$))
\State \textbf{end for}
\State \textbf{return} Decoder($\mathbf{z}$)
\end{algorithmic}
\end{algorithm}

\subsection{Diagnostic intervention framework}\label{sec:A.4}
As shown in Sec.~\ref{sec:intervention} of the main paper, our diagnostic intervention framework 
intervenes only on generations that are flagged as potential failures by the 
Real-time Inspection (RI) module. The intervention is preview-driven: rather 
than applying a fixed correction to every flagged sample, we use the intermediate 
preview signals to decide how to continue the generation, escalating compute 
hierarchically only when needed.

\begin{wrapfigure}{r}{0.47\linewidth}
  \centering
  \includegraphics[width=0.9\linewidth]{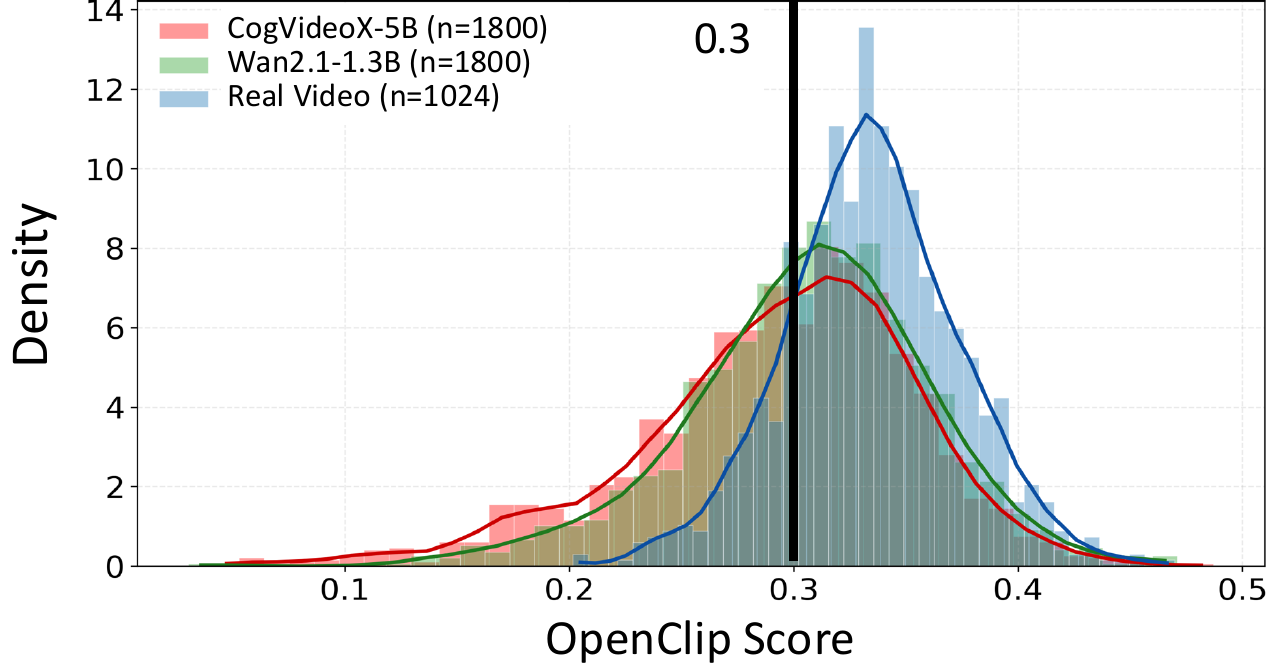}
\caption{\textbf{OpenCLIP score distributions.} We set $\tau = 0.30$ (line) following the procedure of Fig.~\ref{fig:threshold} (main).}
  \label{fig:thres_openclip}
\end{wrapfigure}

\paragraph{Trial 0. Base Generation}
This trial corresponds to the default generation process without any modification.
When the detector predicts $\tilde{s}_{0} \geq \tau$, we simply continue standard 
denoising with the original prompt.

\paragraph{Trial 1. Single-frame Semantic Injection}
Trial~1 targets a specific failure mode in which the model captures the prompt 
in a single frame but fails to extend it into a coherent video. We diagnose this 
from the previews themselves: using the same T2V model with the temporal 
dimension collapsed, we generate a single-frame preview and score its predicted 
alignment $\tilde{s}_{\text{img}}$ with the same dynamic detector. When 
$\tilde{s}_{\text{img}} > \tilde{s}_{0} + \delta$ (with $\delta=0.05$), the 
single frame is better aligned than the video, supporting the temporal-failure 
hypothesis, and we apply single-frame semantic injection.

Algorithm~\ref{alg:single_frame_injection} describes the detailed procedure. For 
simplicity, we denote the single-image intermediate prediction 
$\hat{\mathbf{z}}^{\mathrm{img}}_{0|k_{\mathrm{img}}}$ as $\mathbf{z}^{\mathrm{img}}$.
To make it compatible with video generation, we first expand 
$\mathbf{z}^{\mathrm{img}}$ along the temporal dimension. We then add noise so 
that it matches the latent at the second timestep, where single-frame injection 
begins. During the following two denoising steps, we apply a soft temporal mask 
that assigns a weight of $1$ to the starting frame and linearly decays to $0$ 
over time, blending the injected latent with the model-updated latent. After 
these two steps, the remaining denoising proceeds in the same manner as the 
base generation. Since the latent at $t_1$ is essentially pure noise and carries no 
semantic signal, we start the injection at $k=2$, the earliest step 
that retains usable information from $\mathbf{z}^{\mathrm{img}}$.

\paragraph{Trial 2. Visual Feedback Intervention}
When the single-frame test fails ($\tilde{s}_{\text{img}} \leq \tilde{s}_{0} + 
\delta$), the failure is unlikely to be purely temporal, so Trial~1 is bypassed 
and we proceed directly to Trial~2. We provide the failed video preview and the 
original text prompt to a vision--language model (VLM), and ask it to identify 
the visible mismatch and return an intent-preserving refined prompt with an 
optional negative prompt. Figure~\ref{fig:prompt_template_refinement} presents 
the instruction template used for this step. We then restart generation under 
the refined prompt and monitor it with the same RI module. This trial is 
particularly effective when the failure is caused by missing semantic cues, 
incorrect actions, or weak style descriptions that are difficult to correct 
through latent injection alone.

\paragraph{Trial 2$\rightarrow$1}
If the regenerated run under the refined prompt still satisfies 
$\tilde{s}_{0} < \tau$, the failure likely combines semantic and temporal causes. 
In this case, we apply Trial~1 once on top of the refined prompt before 
terminating, ensuring both the semantic and temporal failure modes are addressed.

\subsection{Threshold calibration for OpenCLIP}\label{sec:A.5}
We measure OpenCLIP scores on the same video set used for ViCLIP threshold calibration in the main paper. Real videos ($1{,}024$ Pexels clips) again concentrate at higher scores than generated videos from CogVideoX-5B and Wan2.1-1.3B (Fig.~\ref{fig:thres_openclip}). Following the same procedure, we set $\tau = 0.30$, which removes the low-score tail of generated distributions and brings the retained samples into the real-video score range, mirroring the $\tau = 0.22$ choice for ViCLIP.

\begin{figure}[t]
  \centering
  \includegraphics[width=0.95\linewidth]{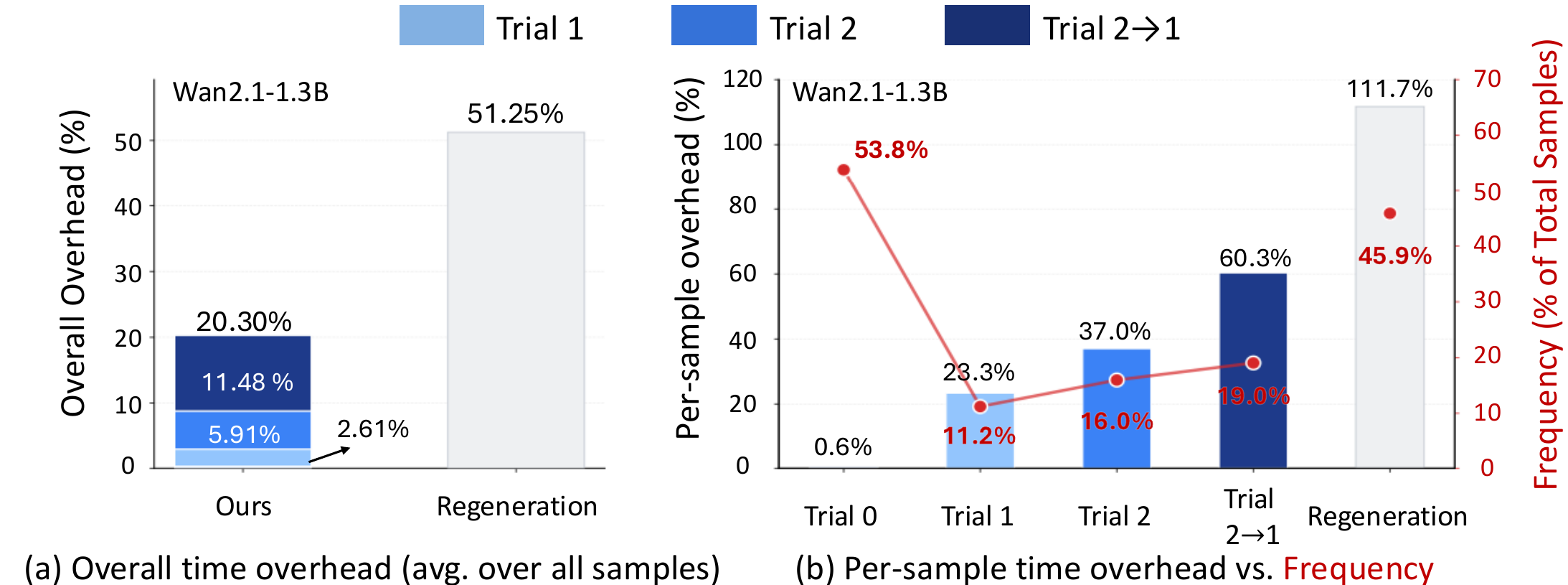}
  \caption{\textbf{Efficiency of selective early intervention on Wan2.1-1.3B.}
  (a) Overall time overhead relative to base generation. Ours incurs only +20.3\% additional time overhead, substantially lower than the regeneration of failure samples (+51.25\%). (b) Per-sample cost increases with deeper trials, yet remains cheaper than regeneration. Since deeper trials are triggered for only a subset of samples (red curve), the overall time overhead stays modest.
  }
  \label{fig:overhead_wan}
\end{figure}

\section{Analysis of time overhead}\label{sec:B}
We provided the efficiency results of selective early intervention for Wan2.1-1.3B, alongside those of CogVideoX-5B shown in Fig.~\ref{fig:cost_stair} of the main paper. Now, we present a detailed breakdown of how the time overhead is computed.

\subsection{Efficiency of selective early intervention on Wan2.1-1.3B}\label{sec:B.1}

Figure~\ref{fig:overhead_wan} provides a detailed breakdown of the efficiency of selective early intervention on Wan2.1-1.3B. As shown in Fig.~\ref{fig:overhead_wan}(b), the per-sample time overhead increases with intervention depth, rising from 23.3\% for Trial~1 to 37.0\% for Trial~2 and 60.3\% for Trial~2$\rightarrow$1. In contrast, naive regeneration incurs a much larger per-sample time overhead of 111.7\% including VLM. 

Despite this increasing per-sample time cost, only a relatively small fraction of samples reach the more expensive stages. Most samples either finish without intervention or are handled by earlier trials, which keeps the average cost low. As a result, the overall time overhead remains modest at 20.30\% over all samples, substantially lower than regeneration at 51.25\%, as shown in Fig.~\ref{fig:overhead_wan}(a).

\subsection{Computation of time overhead}\label{sec:B.2}
We compute time overhead from executions over the full VBench evaluation set of 946 prompts, using two measured quantities: (i) the average runtime of each primitive operation in Table~\ref{tab:overhead_detailed_time}, and (ii) the average number of times each operation is executed in each branch in Table~\ref{tab:overhead_detail}. For each branch, the total runtime is obtained by summing the costs of all executed operations, including video denoising, video preview inspection, single-frame denoising, single-frame preview inspection, and VLM calls.

\begin{table}[t]
\centering
\small
\caption{\textbf{Measured execution time of each primitive operation.}
We report the average runtime of each primitive operation used in our framework, including video denoising, video preview inspection, single-frame denoising, single-frame preview inspection, and VLM calls. 
We also report the average detection step of the dynamic failure detector for video previews and single-frame previews. 
These measured runtimes are used together with the branch-wise execution counts in Table~\ref{tab:overhead_detail} to compute the per-sample and overall time overhead.}
\label{tab:overhead_detailed_time}
\resizebox{\linewidth}{!}{
\begin{tabular}{lccccccc}
\midrule
& \multicolumn{5}{c}{}
& \multicolumn{2}{c}{\textbf{Dynamic failure detector}} \\
& \multicolumn{5}{c}{\textbf{Execution time (s)}}
& \multicolumn{2}{c}{\textbf{average step}} \\
\cmidrule(lr){2-6}
\cmidrule(lr){7-8}
& Video & Video & Single-frame & Single-frame & VLM & Video & Single-frame \\
\textbf{Models} & denoising steps & preview inspections & denoising steps & preview inspections & calls & preview & preview \\
\midrule
CogVideoX-5B
& 4.343
& 0.0392
& 0.405
& 0.0218
& 8.541
& 11.05
& 24.00 \\
Wan2.1-1.3B
& 1.460
& 0.0392
& 0.103
& 0.0218
& 8.541
& 10.33
& 24.00 \\
\bottomrule
\end{tabular}}
\end{table}

\begin{table*}[t]
\centering
\small
\caption{\textbf{Detailed breakdown of execution statistics and time overhead.}
We report the number of denoising steps, inspection operations, and wall-clock runtime for each execution branch. \textit{Per-sample time overhead} measures the additional runtime relative to the original generation time when the branch is executed, while \textit{Overall time overhead} represents the dataset-averaged contribution.}
\label{tab:overhead_detail}
\resizebox{\linewidth}{!}{
\begin{tabular}{lcccccccccc}

\toprule
\multicolumn{11}{c}{\textbf{CogVideoX-5B: original generation time: 217.15 s (50 steps)}} \\
\midrule

& & \multicolumn{5}{c}{\textbf{Execution statistics}}
& \multicolumn{2}{c}{\textbf{Runtime}}
& \multicolumn{2}{c}{\textbf{Time overhead}} \\

\cmidrule(lr){3-7}
\cmidrule(lr){8-9}
\cmidrule(lr){10-11}

\textbf{Branch}
& Samples / Total & \shortstack{Video\\denoising steps}
& \shortstack{Video\\preview inspections}
& \shortstack{Single-frame\\denoising steps}
& \shortstack{Single-frame\\preview inspections}
& \shortstack{VLM\\calls}

& \shortstack{Average\\time (s)}
& \shortstack{Additional\\time (s)}

& \shortstack{Per-sample\\time overhead (\%)}
& \shortstack{Overall\\time overhead (\%)} \\

\midrule

Trial 0 & 562\,/\,946
& 50
& 11.05
& 0
& 0
& 0
& 217.58
& 0.43
& 0.20
& 0.12 \\

Trial 1
& 91\,/\,946 & 60.05
& 11.05
& 24
& 24
& 0
& 271.47
& 54.32
& 25.02
& 2.41 \\

Trial 2
& 125\,/\,946 & 61.05
& 11.05
& 24
& 24
& 1
& 284.36
& 67.21
& 30.95
& 4.09 \\

Trial 2 $\rightarrow$ Trial 1
& 168\,/\,946 & 71.10
& 22.10
& 48
& 48
& 1
& 338.68
& 121.53
& 55.97
& 9.94 \\

\bottomrule

\multicolumn{11}{r}{\textbf{Total overall time overhead: 16.55\%} \quad} \\

\toprule
\multicolumn{11}{c}{\textbf{Wan2.1-1.3B: original generation time: 73.00 s (50 steps)}} \\
\midrule

& & \multicolumn{5}{c}{\textbf{Execution statistics}}
& \multicolumn{2}{c}{\textbf{Runtime}}
& \multicolumn{2}{c}{\textbf{Time overhead}} \\

\cmidrule(lr){3-7}
\cmidrule(lr){8-9}
\cmidrule(lr){10-11}

\textbf{Branch}
& Samples / Total & \shortstack{Video\\denoising steps}
& \shortstack{Video\\preview inspections}
& \shortstack{Single-frame\\denoising steps}
& \shortstack{Single-frame\\preview inspections}
& \shortstack{VLM\\calls}

& \shortstack{Average\\time (s)}
& \shortstack{Additional\\time (s)}

& \shortstack{Per-sample\\time overhead (\%)}
& \shortstack{Overall\\time overhead (\%)} \\

\midrule

Trial 0
& 509\,/\,946 & 50
& 10.33
& 0
& 0
& 0
& 73.40
& 0.40
& 0.55
& 0.30 \\

Trial 1
& 106\,/\,946 & 59.33
& 10.33
& 24
& 24
& 0
& 90.02
& 17.02
& 23.32
& 2.61 \\

Trial 2
& 151\,/\,946 & 60.33
& 10.33
& 24
& 24
& 1
& 100.02
& 27.02
& 37.02
& 5.91 \\

Trial 2 $\rightarrow$ Trial 1
& 180\,/\,946 & 69.66
& 20.66
& 48
& 48
& 1
& 117.04
& 44.04
& 60.34
& 11.48 \\

\bottomrule

\multicolumn{11}{r}{\textbf{Total overall time overhead: 20.30\% \quad}} \\

\end{tabular}}

\end{table*}

We define the original generation time as the runtime of the default 50-step video generation without inspection or intervention. Even Trial~0 incurs a small non-zero overhead due to the preview inspection step. The per-sample time overhead is therefore computed as the additional runtime of each branch relative to the original generation time.

The overall time overhead is computed over the full evaluation set. Specifically, we measure the total runtime of applying our method to all 946 prompts, subtract the total runtime of original generation on the same 946 prompts, and divide the difference by the total original generation time. As a result, deeper branches can have high per-sample overhead, while their contribution to the final average time overhead remains limited because they are triggered less often. For example, on Wan2.1-1.3B, Trial~1 takes 90.02\,s on average, compared with the original generation time of 73.00\,s, resulting in 17.02\,s additional runtime and 23.32\% per-sample time overhead.

The numbers of video preview and single-frame preview inspections are determined by the average detection steps of the dynamic failure detector. For Trial~1 and Trial~2$\rightarrow$1, single-frame semantic injection follows Algorithm~1 that starts from the second denoising step. 
Therefore, the subsequent video denoising process
uses 49 steps instead of 50.

For Wan2.1-14B at 720p with 81 frames, each denoising
step takes 67.052~s for video generation and 1.286~s
for single-frame generation. Applying the same
operation-based runtime accounting yields the additional time overhead of 8.17\%. The lower relative
overhead is due to the greater cost of full-video
denoising in this larger-model regime.

\section{Additional experiments}\label{sec:C}

\begin{figure}[t]
  \centering
  \includegraphics[width=.95\linewidth]{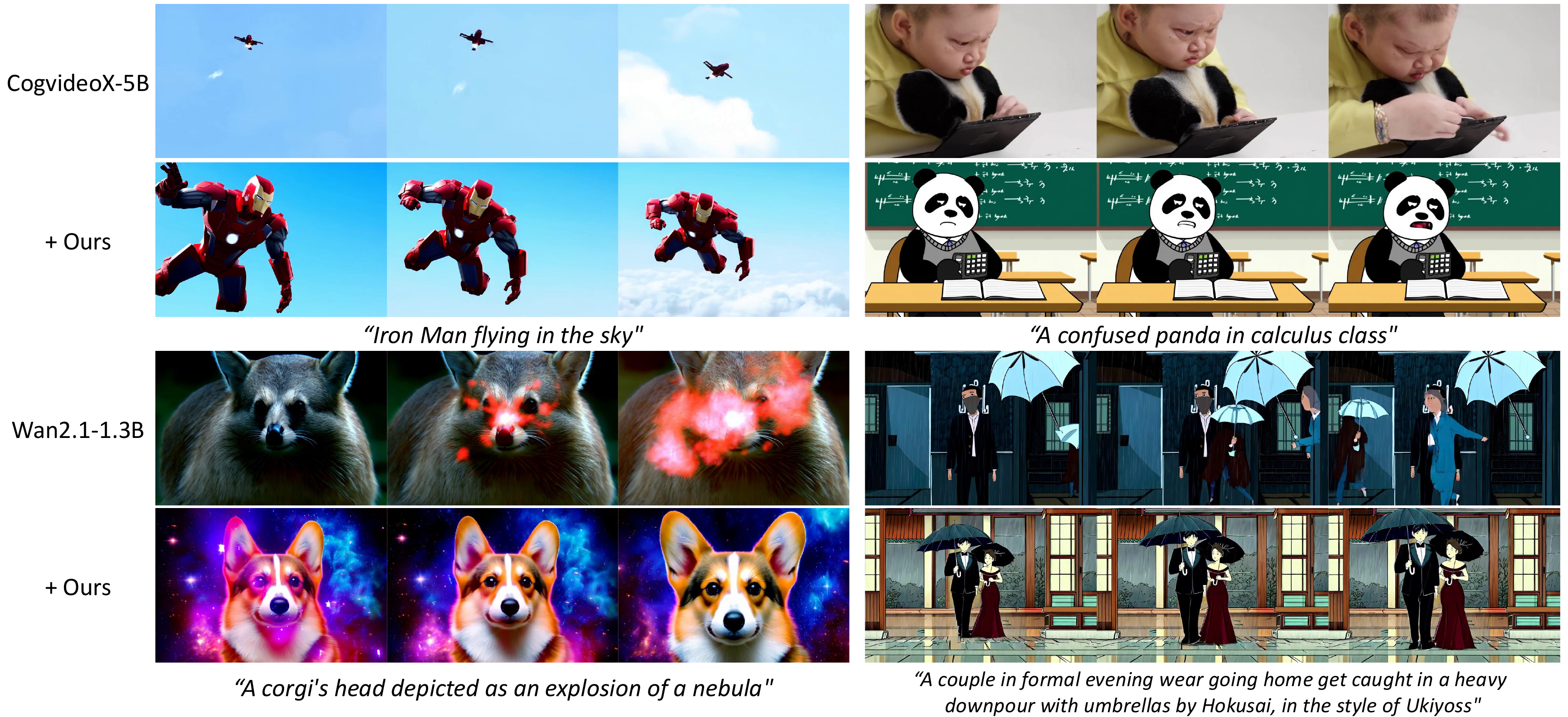}
\caption{\textbf{Qualitative results.} Compared to baseline models, our method 
improves prompt adherence and visual fidelity by correcting early-failure 
generations. Top two rows: CogVideoX-5B. Bottom two rows: Wan2.1-1.3B.}
  \label{fig:qualitative}
\end{figure}

\begin{figure}[t]
  \centering
  \includegraphics[width=0.95\linewidth]{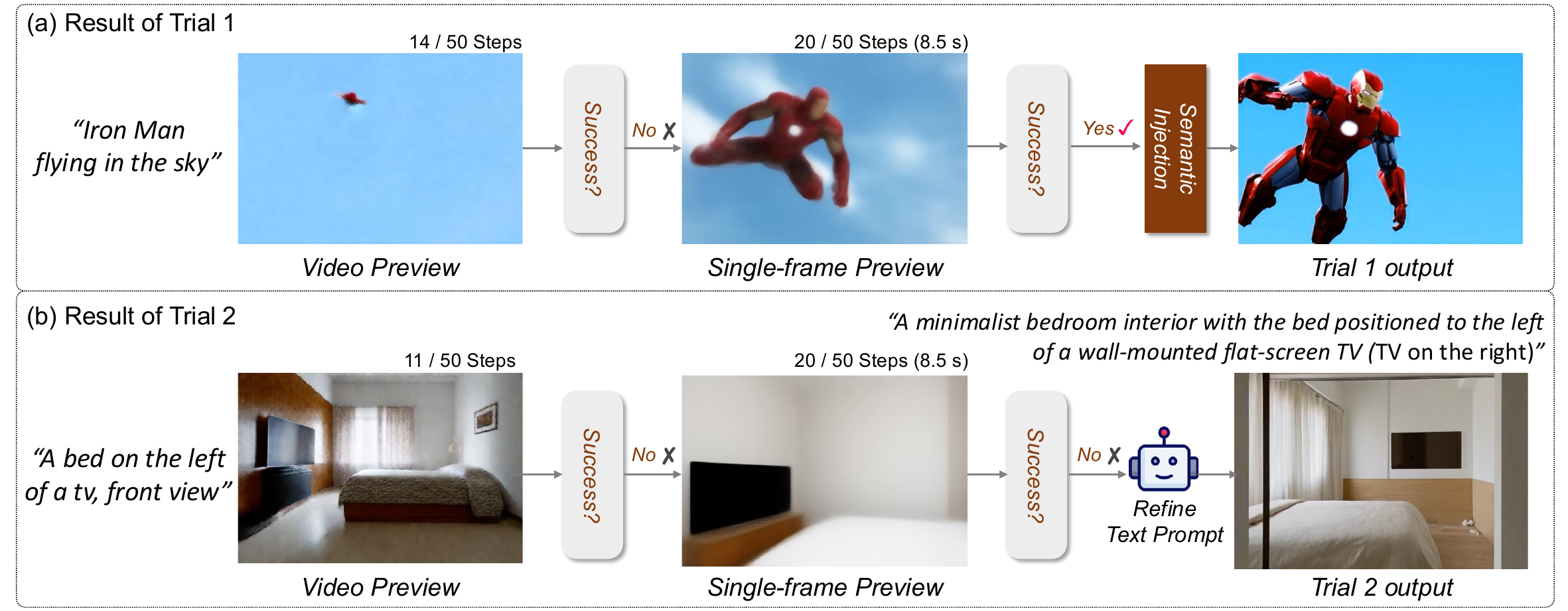}
\caption{\textbf{Case visualization of diagnostic interventions.} 
\textbf{Top:} although the early video preview poorly reflects the prompt, the 
single-frame preview captures the semantics. The pipeline applies 
\textit{single-frame semantic injection} (Trial~1), correcting the subject 
appearance without invoking a VLM. \textbf{Bottom:} both the video and single-frame previews place the TV on the left, violating the intended relation that the bed
should be to the left of the TV. This demonstrates persistent prompt misalignment beyond 
temporal causes. The pipeline therefore invokes \textit{Visual Feedback 
Intervention} (Trial~2), where a VLM refines the prompt based on the failed 
preview before regeneration.}
\label{fig:intervene_sample}
\end{figure}

\subsection{Qualitative results of our pipeline}\label{sec:C.2}
Figure~\ref{fig:qualitative} presents qualitative comparisons between baseline 
generations with and without our method on samples predicted as failures. 
Baseline outputs sometimes drift toward semantically misaligned content or 
exhibit weak subject depiction, whereas our pipeline mitigates these failure 
modes through early detection and selective intervention.

Figure~\ref{fig:intervene_sample} illustrates how the intervention framework 
performs hierarchical diagnostic trials. The Real-time Inspection (RI) module 
analyzes intermediate previews and predicts potential failures early in the 
denoising process, triggering targeted interventions with an early-exit 
strategy. In the top example, the early video preview poorly reflects the 
prompt while the single-frame preview captures the semantics, so the pipeline 
applies \textit{single-frame semantic injection} (Trial~1), which corrects the 
subject appearance without invoking a VLM. In contrast, the bottom example 
shows a persistent prompt misalignment: both the video preview and Trial~1 
fail to satisfy the intended spatial relationship (the TV appearing on the 
left), indicating a failure beyond purely temporal causes. The pipeline 
therefore escalates to \textit{Visual Feedback Intervention} (Trial~2), in 
which a VLM analyzes the failed preview and refines the prompt before 
regeneration.  Additional qualitative examples of Trial~1 and Trial~2 are 
provided in Fig.~\ref{fig:trial1_qual} and Fig.~\ref{fig:trial2_qual}, 
respectively.

\begin{table}[t]
\centering
\caption{\textbf{Five-seed evaluation on VBench.} We re-run both base models in the Original prompt regime under VBench's official five-seed protocol. Results closely match the single-seed numbers in the main paper (Table~\ref{tab:vbench_regen}), confirming that our reported gains are robust to seed choice.}
\label{tab:vbench_multiseed}
\resizebox{0.9\linewidth}{!}{
\begin{tabular}{@{} l ccc ccc @{}}
\toprule
\multirow{2}[2]{*}{\textbf{Method}} & \multicolumn{3}{c}{\textbf{CogVideoX-5B}} & \multicolumn{3}{c}{\textbf{Wan2.1-1.3B}} \\
\cmidrule(lr){2-4} \cmidrule(lr){5-7}
& \textbf{Final} & \textbf{Quality} & \textbf{Semantic} & \textbf{Final} & \textbf{Quality} & \textbf{Semantic} \\
\midrule
Original prompt (baseline) & 0.776 & 0.803 & 0.670 & 0.787 & 0.813 & 0.686 \\
\quad + Ours & \textbf{0.796} & \textbf{0.811} & \textbf{0.736} & \textbf{0.806} & \textbf{0.820} & \textbf{0.750} \\
\bottomrule
\end{tabular}
}
\end{table}

\subsection{Multi-seed evaluation on VBench}\label{sec:C.3}
To complement the main paper results, we additionally evaluate under the VBench five-seed protocol on both base models in the Original prompt regime. The multi-seed numbers (Table~\ref{tab:vbench_multiseed}) closely match the main paper: baseline Final scores differ by at most $0.003$, and the gains from our intervention are nearly identical ($+0.020$ vs.\ $+0.022$ on CogVideoX-5B Final; $+0.019$ vs.\ $+0.016$ on Wan2.1-1.3B Final). The qualitative pattern is preserved across all dimensions. These results corroborate the main findings and confirm that the gains are not seed-specific.

\begin{wrapfigure}{r}{0.47\linewidth}
  \centering
  \includegraphics[width=0.9\linewidth]{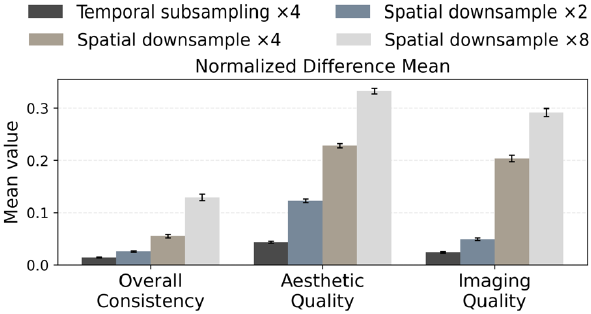}
    \caption{\textbf{Sensitivity of video quality metrics to temporal subsampling and spatial downsampling.}
    Compared to temporal subsampling, spatial downsampling causes larger deviations in VBench scores from the original videos.}
  \label{fig:temporal_resolution}
\end{wrapfigure}

\subsection{Overhead comparison with test-time scaling methods}\label{sec:C.4}
Recent test-time scaling methods for video generation~\citep{liu2025video,oshima2025inference} share our goal of improving inference-time quality but take a fundamentally different approach: they search over multiple candidate trajectories at every step, whereas we monitor a single trajectory and intervene selectively.

Direct comparison is not straightforward: while Video-T1~\citep{liu2025video} reports CogVideoX-5B results in its paper, the released code does not currently support CogVideoX-5B. DLBS~\citep{oshima2025inference}, in turn, reports compute in NFE rather than wall-clock time, following the convention of inference-time scaling literature~\citep{ma2025scaling, liu2025video}.

We therefore report two complementary indicators of computational cost. First, on VBench, DLBS and its lookahead variant DLBS-LA report 1200 to 2400 NFE and 2500 to 5000 NFE, respectively, compared with 50 NFE for standard generation, corresponding to 24 to 48$\times$ and 50 to 100$\times$ more NFEs. Second, to obtain a concrete wall-clock reference, we ran the official DLBS \texttt{very\_high} configuration ($K{=}2$, $B{=}4$, 50 DDIM steps, 49 frames, bf16) on CogVideoX-5B, on a single A100; this run takes 4 hours and 32 minutes per video.

These numbers are not intended as a strictly controlled benchmark, since NFE does not always translate linearly to wall-clock time under parallel batching or implementation-specific optimizations. Nevertheless, the order-of-magnitude gap relative to our measured 16.55\% (CogVideoX-5B) and 20.30\% (Wan2.1-1.3B) overhead suggests that multi-trajectory test-time scaling and our selective single-trajectory intervention operate in qualitatively different cost regimes. We therefore use this comparison only to contextualize the computational regime, not to claim a head-to-head runtime or quality advantage under matched implementations.

\subsection{Spatiotemporal sensitivity of VBench metrics}\label{sec:C.5}
This section provides the empirical analysis behind the L2R converter's 
asymmetric spatiotemporal decoding design described in Sec.~\ref{sec:early_detection} of the main 
paper. Specifically, the L2R converter fully restores the $8\times$ spatial 
compression of the latent to preserve high-fidelity visual structures, while 
retaining the $4\times$ temporal compression inherent to the latent space 
without temporal upsampling. The rationale is that VBench metrics, which 
serve as our inspection signal, are substantially more sensitive to spatial 
than temporal resolution.

To verify this, we collect 1{,}024 real videos from Pexels~\cite{Pexels}, 
apply spatial downsampling and temporal subsampling at various rates, and 
measure the changes in Overall Consistency, Aesthetic Quality, and Imaging 
Quality~\cite{huang2024vbench} relative to the original videos. As shown in 
Fig.~\ref{fig:temporal_resolution}, the metrics are substantially more 
sensitive to spatial resolution: even mild spatial downsampling ($\times 2$) 
yields a larger normalized MAE than temporal subsampling ($\times 4$). This 
asymmetry justifies our design choice to spend the L2R converter's modeling 
capacity on spatial restoration while leaving the temporal axis at the latent's 
frame count, keeping the converter compact (0.059\,M parameters) without 
sacrificing the inspection signal.

\begin{figure}[t]
  \centering
  \includegraphics[width=0.85\linewidth]{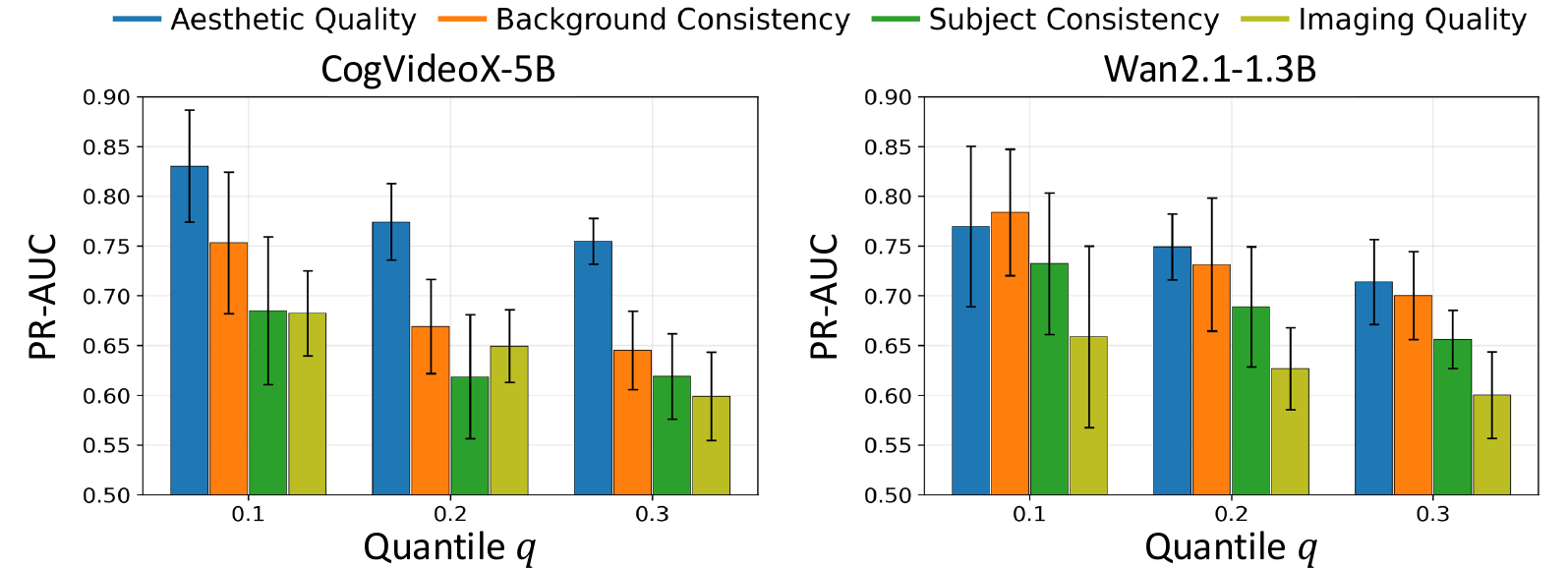}
  \caption{\textbf{Overall Consistency as a proxy for other VBench metrics.}
Using all VBench prompts, we measure PR-AUC for predicting whether a sample falls into the top-$q$ or bottom-$q$ group of each target metric ($q\in\{0.1,0.2,0.3\}$) using Overall Consistency. All scores exceed the random baseline (0.5), indicating that Overall Consistency captures quality variations reflected by other metrics.}
  \label{fig:overall_consistency_auc}
\end{figure}

\begin{table}[t]
\centering
\caption{\textbf{Comparison with a simple Best-of-3 baseline.}
For \textit{Best-of-3}, we generate three videos per prompt from the original prompt using different random seeds, score them with our alignment scorer (ViCLIP), and select the sample with the highest semantic alignment score. Since this requires three full generations, it incurs 200\% additional time overhead relative to single-sample generation.}
\label{tab:bestof3}
\resizebox{\linewidth}{!}{
\begin{tabular}{lcccccccc}
\toprule
& \multicolumn{4}{c}{CogVideoX-5B} & \multicolumn{4}{c}{Wan2.1-1.3B} \\
\cmidrule(lr){2-5} \cmidrule(lr){6-9}
Method & Final & Quality & Semantic & Overhead (\%) & Final & Quality & Semantic & Overhead (\%) \\
\midrule
Original prompt (baseline) & 0.778 & 0.803 & 0.680 & 0.00 & 0.789 & 0.811 & 0.700 & 0.00 \\
Best-of-3 & 0.787 & 0.802 & 0.728 & 200.00 & 0.796 & 0.812 & 0.730 & 200.00 \\
Original prompt + Ours & \textbf{0.800} & \textbf{0.816} & \textbf{0.735} & \textbf{16.55} & \textbf{0.805} & \textbf{0.817} & \textbf{0.756} & \textbf{20.30} \\
\bottomrule
\end{tabular}
}
\end{table}

\subsection{Overall Consistency as a proxy for video quality}\label{sec:C.6}
This section provides the empirical analysis behind the choice of 
\textit{Overall Consistency} as the single alignment signal used by the RI 
module, briefly summarized in Sec.~\ref{sec:exp-why} of the main paper. To fit the per-step 
latency budget of 19.5\,ms, the RI module relies on a single alignment-based 
score; we examine here whether this single signal serves as a reliable proxy 
for broader video quality, that is, whether it can discriminate quality 
changes captured by other VBench metrics.

To this end, we generate videos using the full set of 946 VBench prompts and 
compute multiple VBench scores. For each target metric, we form a balanced 
binary classification task by selecting samples from the top-$q$ and bottom-$q$ 
quantiles ($q \in \{0.1, 0.2, 0.3\}$), and predict whether a sample belongs to 
the top or bottom group using Overall Consistency. We report PR-AUC with 
10-fold cross-validation; since the subsets are balanced, a random classifier 
yields PR-AUC $\approx 0.5$.

As shown in Fig.~\ref{fig:overall_consistency_auc}, Overall Consistency 
consistently achieves notable AUC scores across all evaluated dimensions. 
Performance is generally highest at $q=0.1$, indicating stronger separability 
when focusing on more extreme cases. The signal is particularly predictive of 
Aesthetic Quality and Background Consistency, while remaining informative for 
Subject Consistency and Imaging Quality. These results justify our use of 
Overall Consistency as a single, low-cost inspection signal: it captures 
quality variations across multiple VBench dimensions, which is consistent with 
the broad per-dimension improvements reported in Table~\ref{tab:vbench_per_dim}.

\subsection{Comparison with a best-of-3 baseline}\label{sec:C.7}

To further contextualize the efficiency--quality trade-off of our method, we compare it with a simple inference-time scaling baseline based on multi-seed sampling. For each prompt, we generate three videos from the original prompt using three different random seeds, compute the semantic alignment score of each sample using our alignment scorer (ViCLIP), and select the sample with the highest score to form the final evaluation set. We denote this baseline as \textit{Best-of-3}. Since it requires three full generations per prompt, it incurs 200\% additional time overhead relative to single-sample generation.

Table~\ref{tab:bestof3} compares \textit{Best-of-3} with the original prompt baseline and our method on VBench. On CogVideoX-5B, \textit{Best-of-3} improves the semantic score over the original baseline, but our method achieves a higher final score (0.800 vs.\ 0.787) and higher quality score (0.816 vs.\ 0.802) with only 16.55\% overhead, compared with 200\% for \textit{Best-of-3}. On Wan2.1-1.3B, \textit{Best-of-3} yields competitive final and semantic scores, but also requires substantially larger inference cost. These results suggest that, even compared with a simple multi-seed selection strategy using the same alignment scorer, our selective early intervention provides a substantially more favorable efficiency--quality trade-off.

\begin{table}[t]
\centering
\caption{\textbf{Effect of the success and failure threshold.}
We compare the original prompt baseline with our selective early intervention under different failure thresholds $\tau$. Higher thresholds make the detector more conservative, increasing the number of samples that receive intervention. As a result, VBench scores may improve further, but at the cost of substantially higher time overhead.}
\label{tab:threshold_ablation}
\resizebox{\linewidth}{!}{
\begin{tabular}{lcccccccc}
\toprule
& \multicolumn{4}{c}{CogVideoX-5B} & \multicolumn{4}{c}{Wan2.1-1.3B} \\
\cmidrule(lr){2-5} \cmidrule(lr){6-9}
Method & Final & Quality & Semantic & Overhead (\%) & Final & Quality & Semantic & Overhead (\%) \\
\midrule
Original prompt (baseline) & 0.778 & 0.803 & 0.680 & 0.00 & 0.789 & 0.811 & 0.700 & 0.00 \\
Original prompt + Ours ($\tau=0.22$) & 0.800 & 0.816 & 0.735 & 16.55 & 0.805 & 0.817 & 0.756 & 20.30 \\
Original prompt + Ours ($\tau=0.26$) & \textbf{0.807} & 0.824 & \textbf{0.740} & 32.61 & 0.807 & 0.823 & 0.743 & 35.96 \\
Original prompt + Ours ($\tau=0.30$) & 0.806 & \textbf{0.827} & 0.725 & 39.47 & \textbf{0.812} & \textbf{0.826} & 0.756 & 42.77 \\
\bottomrule
\end{tabular}
}
\end{table}

\begin{figure}[t]
  \centering
  \includegraphics[width=\linewidth]{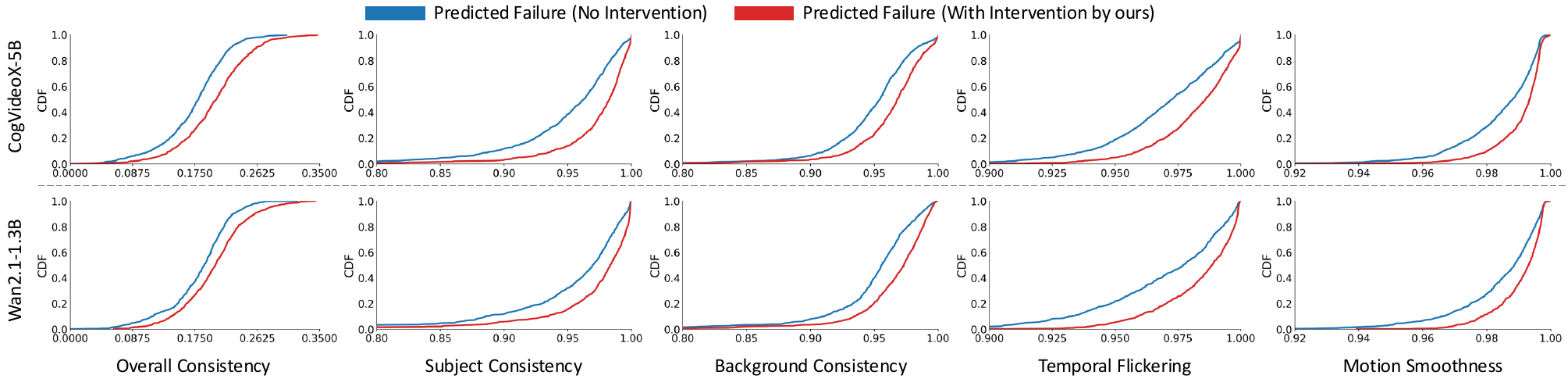}
\caption{\textbf{Intervention improves detected failure cases.}
CDFs of VBench metrics for samples predicted to fail and
forced to complete without intervention, compared with
our intervention. Despite using Overall Consistency for
detection, our method shifts multiple VBench metrics rightward.}
  \label{fig:success_vs_failure_metric}
\end{figure}

\subsection{Threshold sensitivity of early failure detection}\label{sec:C.8}

We study the sensitivity of our method to the failure threshold $\tau$ used by the dynamic failure detector. As shown in Table~\ref{tab:threshold_ablation}, increasing $\tau$ causes more samples to be treated as likely failures and therefore receive intervention. This can further improve VBench scores, but it also leads to substantially larger time overhead on both CogVideoX-5B and Wan2.1-1.3B.

This trend is consistent with the observation in Fig.~\ref{fig:threshold} of the main paper. The distribution of Overall Consistency for generated videos is already shifted to the left of the real-video distribution. As a result, increasing the threshold moves a larger portion of generated samples into the failure region, which raises the intervention trigger frequency. Consequently, higher thresholds can improve aggregate scores by intervening on more borderline samples, but they also increase inference cost significantly.

Consistent with this observation, 
increasing $\tau$  generally improves VBench scores up to a point, while substantially increasing time overhead.
These results show that the failure threshold acts as a practical control knob for the quality--efficiency trade-off of selective early intervention.

\subsection{Effectiveness of intervention on failure cases}\label{sec:C.9}
We evaluate the impact of our intervention on samples flagged as early failures 
by the Real-time Inspection (RI) module. Because the dynamic failure detector 
operates per-model, the number of flagged samples differs slightly across 
generators. Out of the 1{,}800 prompts from VBench and Panda70M, the detector 
flags 872 samples as failures for CogVideoX-5B and 811 for Wan2.1-1.3B.
Figure~\ref{fig:success_vs_failure_metric} compares the VBench metric CDFs of 
these flagged samples with and without our intervention pipeline. Although early 
failure detection is driven by a single signal (Overall Consistency), the 
intervention consistently improves multiple VBench dimensions, with clear 
rightward shifts in Overall Consistency, Subject Consistency, Background Consistency, 
Temporal Flickering, and Motion Smoothness across both CogVideoX-5B and 
Wan2.1-1.3B. This indicates that correcting failures detected through a single 
alignment-based signal yields broad quality gains beyond alignment alone.

\begin{figure}[t]
  \centering
  \includegraphics[width=0.95\linewidth]{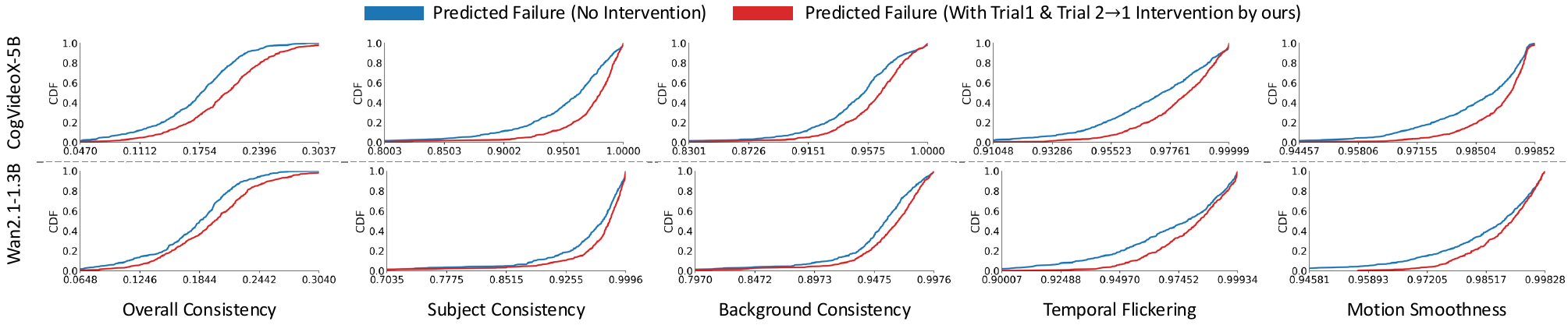}
\caption{\textbf{Effect of the single-frame-based intervention family on detected failure cases.}
We compare the CDFs of VBench metrics between predicted failure samples generated to completion without intervention and those corrected by the single-frame-based intervention family (Trial~1 and Trial~2$\rightarrow$1). Across both CogVideoX-5B and Wan2.1-1.3B, the intervention shifts the distributions toward better scores over multiple VBench dimensions.}
  \label{fig:singleframe_family_cdf}
\end{figure}

\begin{figure}[t]
  \centering
  \includegraphics[width=0.95\linewidth]{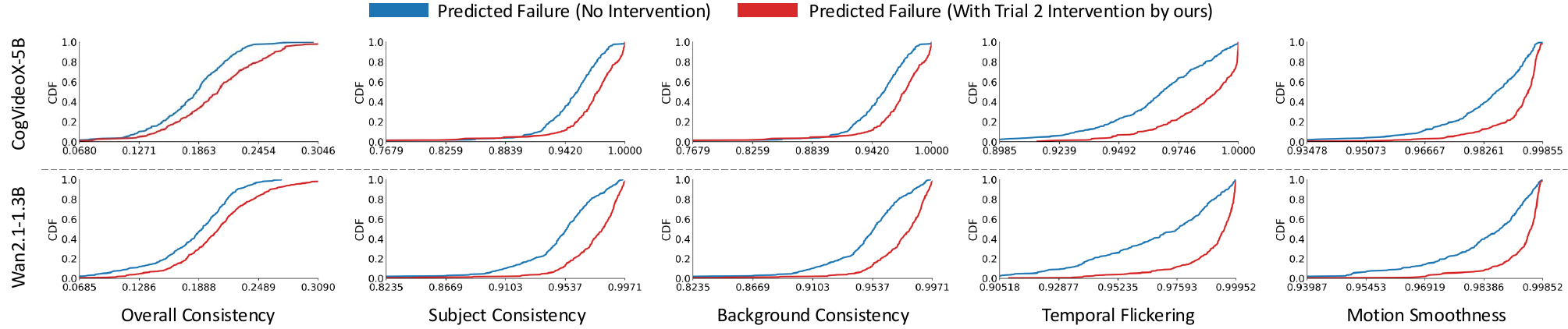}
   \caption{\textbf{Effect of Visual Feedback Intervention on detected failure cases.}
We compare the CDFs of VBench metrics between predicted-failure samples generated to completion without intervention and those corrected by Visual Feedback Intervention (Trial~2). Across both CogVideoX-5B and Wan2.1-1.3B, this branch shifts the distributions toward better scores over multiple VBench dimensions.}
  \label{fig:refined_family_cdf}
\end{figure}

\subsection{Disaggregated analysis of intervention effects}\label{sec:C.10}
The aggregate CDF shifts in Sec.~\ref{sec:C.9} combine the effects of all 
intervention branches. To isolate where the gains come from, we further decompose 
the results into two groups based on the final-stage correction applied: the 
\emph{single-frame-based intervention family}, which includes Trial~1 and 
Trial~2$\rightarrow$1, and \emph{Visual Feedback Intervention} (Trial~2).

\paragraph{Single-frame-based intervention family}
Figure~\ref{fig:singleframe_family_cdf} analyzes the family of branches that 
apply single-frame semantic injection as the final correction step. We group 
Trial~1 and Trial~2$\rightarrow$1 together, since both branches share the same 
final-stage intervention. Compared with predicted-failure samples generated to 
completion without intervention, the corrected samples show consistent rightward 
shifts in the CDFs across multiple VBench dimensions on both CogVideoX-5B and 
Wan2.1-1.3B. This experiment includes 629 samples on CogVideoX-5B and 505 
samples on Wan2.1-1.3B, out of 1{,}800 prompts.

\paragraph{Visual Feedback Intervention}
Figure~\ref{fig:refined_family_cdf} analyzes the effect of Trial~2, which 
invokes a VLM as a visual feedback reasoner over the failed preview to refine 
the prompt before regenerating. As in the main paper, we compare 
predicted-failure samples without intervention against those corrected by this 
branch. The resulting CDF shifts show that Visual Feedback Intervention also 
improves multiple VBench dimensions, complementing the single-frame-based 
family. This experiment includes 243 samples on CogVideoX-5B and 306 samples on 
Wan2.1-1.3B, out of 1{,}800 prompts.

\section{Additional qualitative results}\label{sec:D}
\begin{figure}[t]
  \centering
  \includegraphics[width=.95\linewidth]{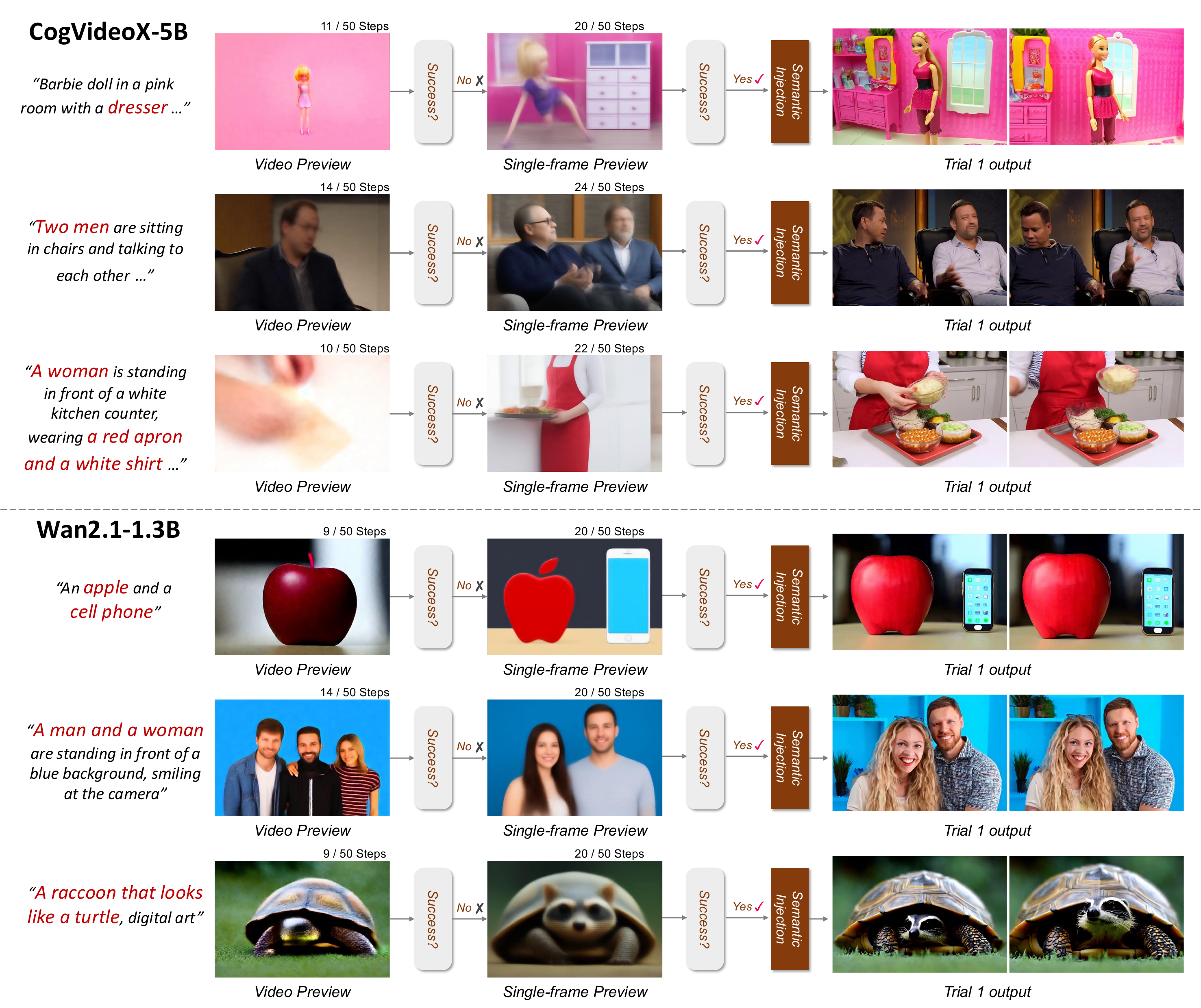}
\caption{\textbf{Additional qualitative results of Trial~1 (single-frame semantic injection).}
We present additional qualitative results from CogVideoX-5B (top) and Wan2.1-1.3B (bottom). For generations predicted as failures based on early video previews, Trial~1 generates a single-frame preview using the same T2V model with the temporal dimension collapsed. When the single-frame preview yields a higher alignment score than the video preview ($\tilde{s}_{\text{img}} > \tilde{s}_{0} + \delta$), we reuse its intermediate prediction as a semantic anchor and inject it into the video latent over the first denoising steps. These examples show that Trial~1 improves prompt fidelity by recovering missing objects, attributes, and subject composition across diverse models and failure cases.}
\label{fig:trial1_qual}
\end{figure}

\begin{figure}[t]
  \centering
  \includegraphics[width=.95\linewidth]{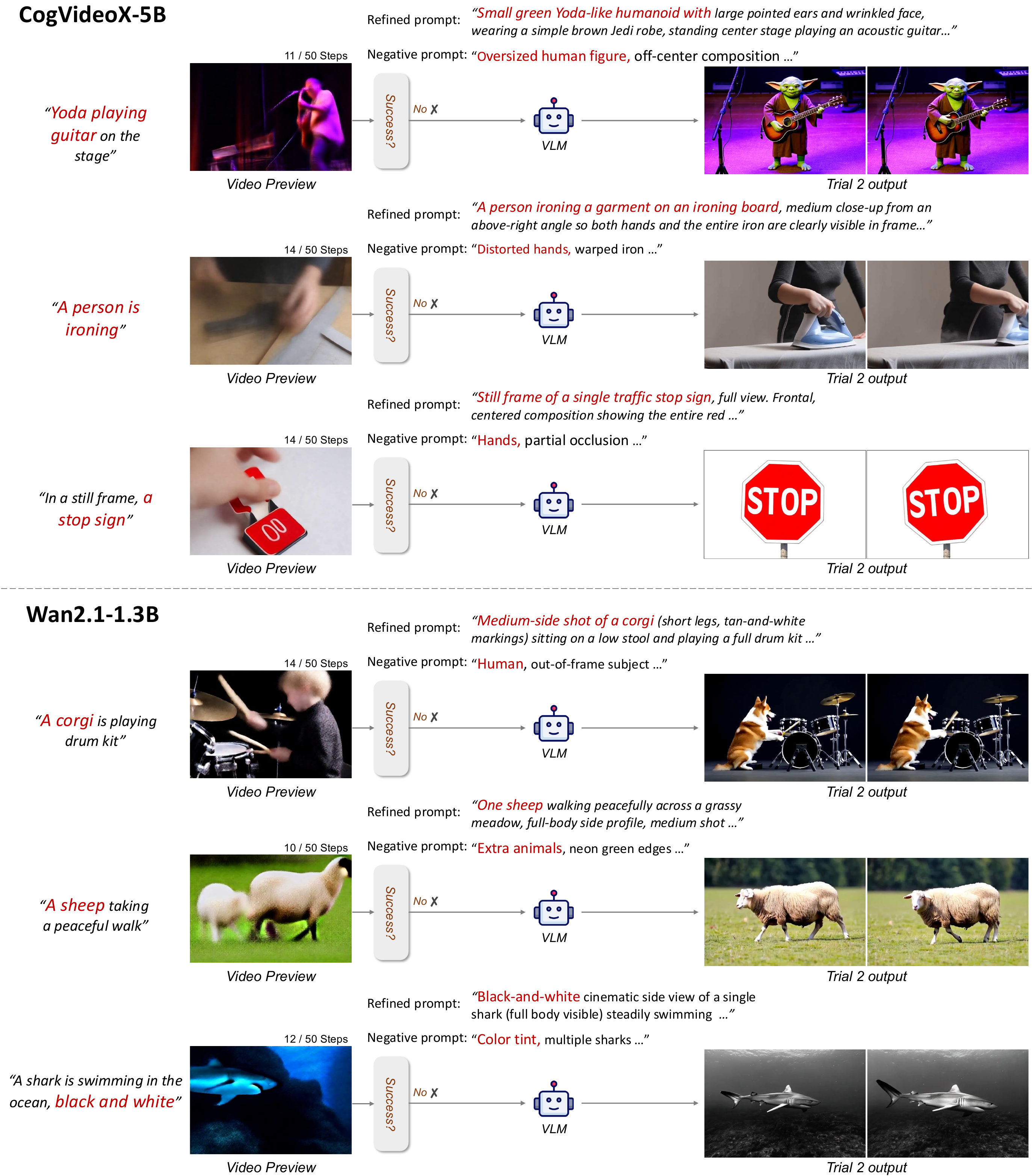}
\caption{\textbf{Additional qualitative results of Trial~2 (Visual Feedback Intervention).}
We present additional qualitative results from CogVideoX-5B (top) and Wan2.1-1.3B (bottom). For generations predicted as failures based on early video previews, Trial~2 invokes a vision--language model (VLM) as a visual feedback reasoner over the failed preview, returning a refined prompt and an optional negative prompt. The refined prompt strengthens the desired semantics, while the negative prompt suppresses undesired content or artifacts. These examples show that Trial~2 improves prompt fidelity by correcting subject identity, object count, composition, and visual attributes across diverse models and failure cases.}
  \label{fig:trial2_qual}
\end{figure}

\begin{figure}[t]
  \centering
  \includegraphics[width=.95\linewidth]{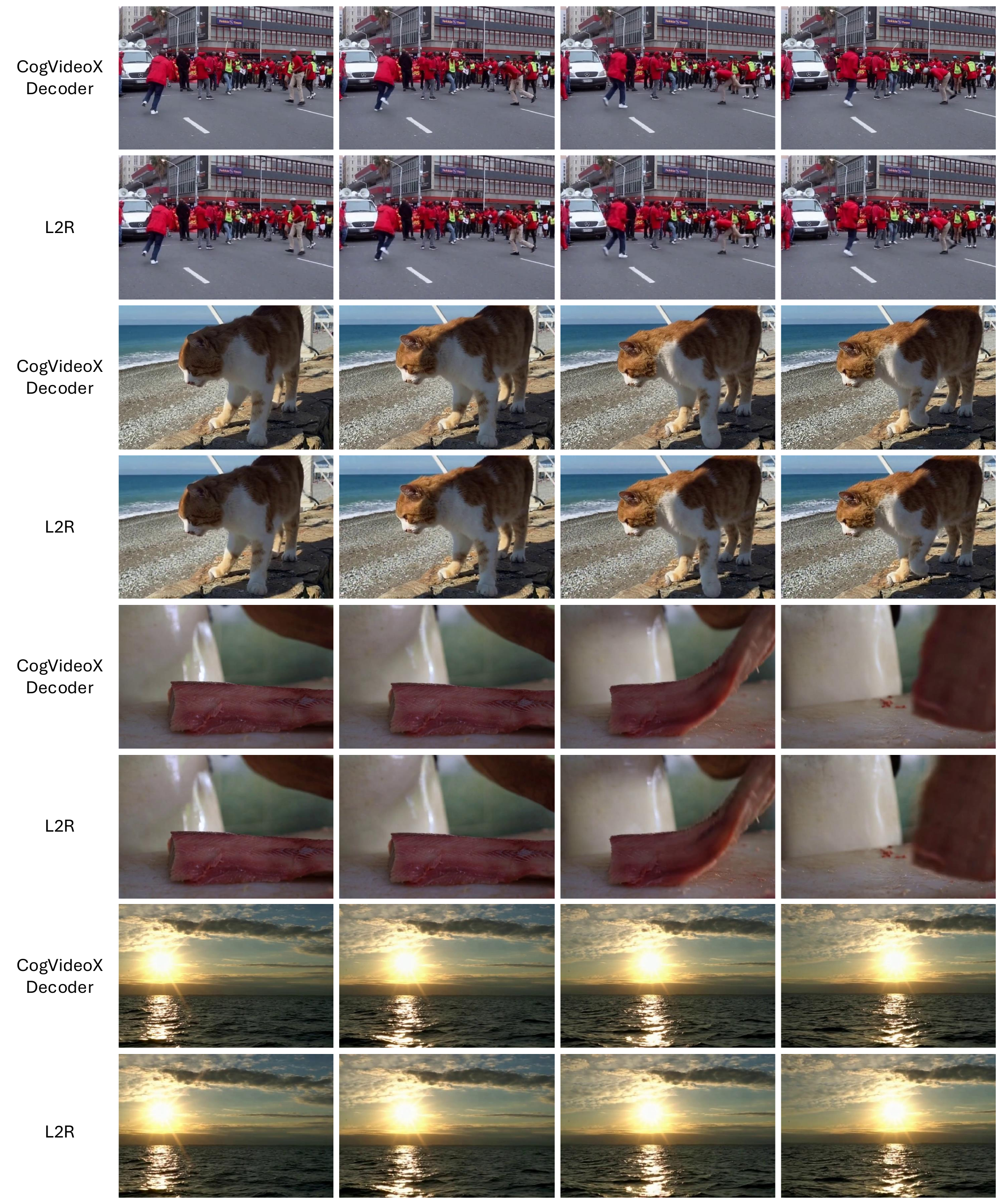}
\caption{\textbf{Qualitative comparison between the native CogVideoX decoder and L2R converter.} We compare the reconstruction results between two methods.
Across diverse scenes and multiple frames, L2R produces lightweight video previews that preserve the key semantic content and visual structure of the native CogVideoX decoder. These previews are sufficiently informative for real-time inspection and failure detection.}
\label{fig:L2R_cogvideox}
\end{figure}

\begin{figure}[t]
  \centering
  \includegraphics[width=.95\linewidth]{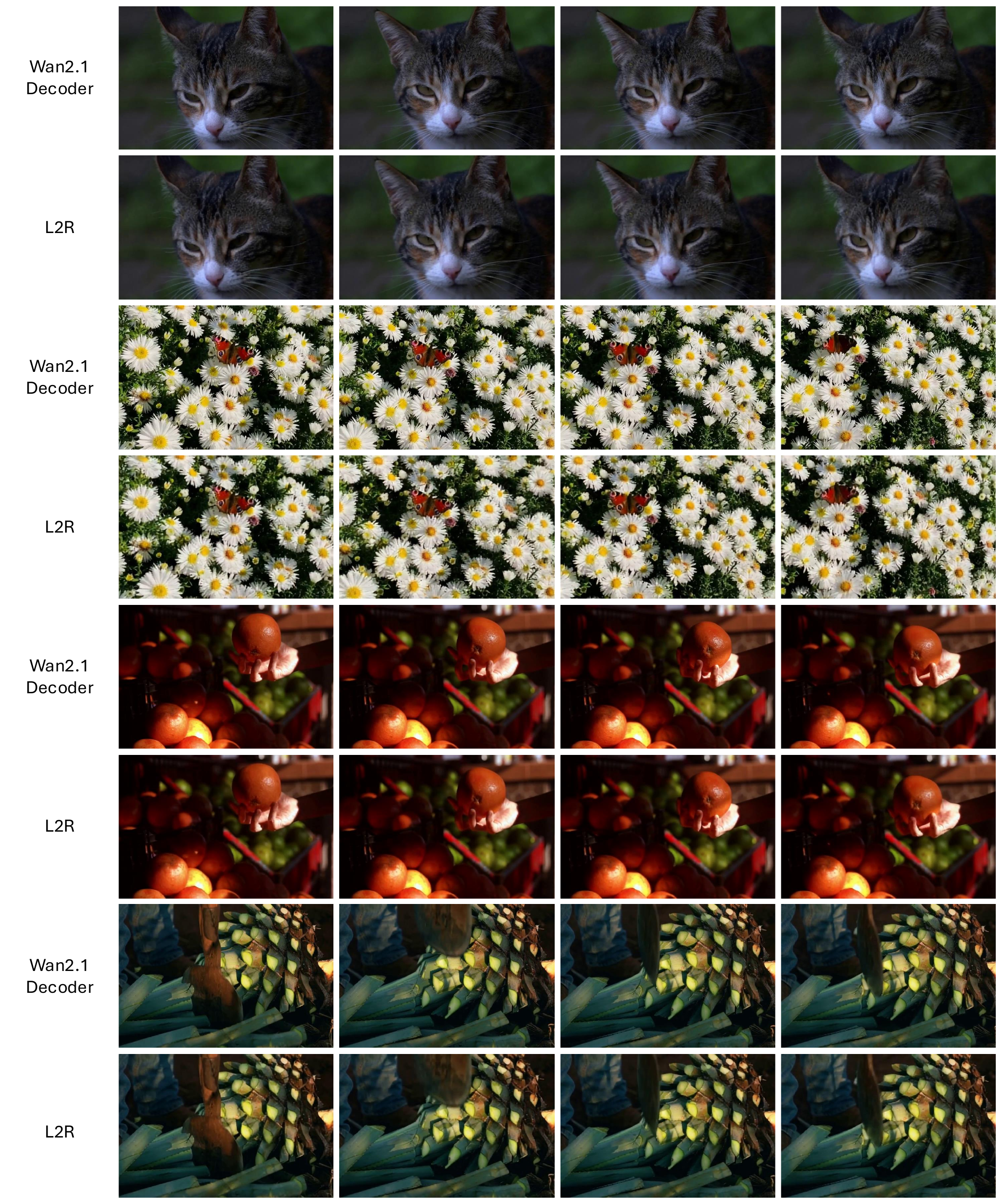}
\caption{\textbf{Qualitative comparison between the native Wan2.1 decoder and L2R converter.} We compare the reconstruction results between two methods. 
Across diverse scenes and multiple frames, L2R produces lightweight video previews that preserve the key semantic content and visual structure of the native Wan2.1 decoder. These previews are sufficiently informative for real-time inspection and failure detection.}
\label{fig:L2R_wan}
\end{figure}

In this section, we provide additional qualitative results for the proposed 
intervention framework. Specifically, we present representative examples of 
\textbf{Trial~1} (\emph{single-frame semantic injection}) and \textbf{Trial~2} 
(\emph{Visual Feedback Intervention}) to illustrate how the proposed method 
corrects early-failure trajectories during generation in Fig.~\ref{fig:trial1_qual} 
and Fig.~\ref{fig:trial2_qual}, respectively. We also provide qualitative 
comparisons between the native decoders of CogVideoX-5B and Wan2.1-1.3B and 
the corresponding outputs of the proposed Latent-to-RGB (L2R) converter in 
Fig.~\ref{fig:L2R_cogvideox} and Fig.~\ref{fig:L2R_wan}, respectively, 
showing that L2R preserves the key semantic content and visual structure needed for real-time inspection.

\clearpage

\begin{table}[t]
\centering
\caption{\textbf{All VBench results across all dimensions.}
We report all individual VBench dimension scores, along with the aggregated Semantic, Quality, and Final scores, for the original prompt baseline and our method on CogVideoX-5B and Wan2.1-1.3B. Overall, our method improves most VBench dimensions on both models, indicating that selective early intervention benefits not only alignment-related metrics but also broader perceptual and temporal aspects of video quality.}
\label{tab:vbench_per_dim}
\resizebox{0.95\linewidth}{!}{
\begin{tabular}{lcccc}
\toprule
\textbf{Dimension} & \multicolumn{2}{c}{\textbf{CogVideoX-5B}} & \multicolumn{2}{c}{\textbf{Wan2.1-1.3B}} \\
\cmidrule(lr){2-3}\cmidrule(lr){4-5}
& \textbf{Original Prompt} & \textbf{+ Ours} & \textbf{Original Prompt} & \textbf{+ Ours} \\
\midrule
Subject Consistency     & 0.957 & \textbf{0.967} & 0.949 & \textbf{0.964} \\
Background Consistency  & 0.953 & \textbf{0.964} & 0.959 & \textbf{0.972} \\
Temporal Flickering     & 0.969 & \textbf{0.973} & 0.991 & \textbf{0.992} \\
Motion Smoothness       & 0.981 & \textbf{0.985} & 0.970 & \textbf{0.979} \\
Dynamic Degree          & \textbf{0.542} & \textbf{0.542} & \textbf{0.542}  & 0.403 \\
Aesthetic Quality       & 0.592 & \textbf{0.600} & 0.590 & \textbf{0.610} \\
Imaging Quality         & 0.610 & \textbf{0.634} & 0.644  & \textbf{0.668} \\
Object Class            & 0.815 & \textbf{0.866} & 0.828 & \textbf{0.900} \\
Multiple Objects        & 0.495 & \textbf{0.656} & 0.630 & \textbf{0.697} \\
Human Action            & 0.870 & \textbf{0.960} & 0.770 & \textbf{0.910} \\
Color                   & 0.843 & \textbf{0.859} & 0.934 & \textbf{0.943} \\
Spatial Relationship    & 0.450 & \textbf{0.498} & 0.687 & \textbf{0.705} \\
Scene                   & 0.393 &\textbf{0.472} & 0.289 & \textbf{0.398} \\
Appearance Style        & 0.229 & \textbf{0.231} & 0.222 & \textbf{0.223} \\
Temporal Style          & 0.238 & \textbf{0.244} & 0.238 & \textbf{0.243} \\
Overall Consistency     & 0.258 & \textbf{0.267} & 0.243 & \textbf{0.256} \\
\midrule
\textbf{Semantic Score} & 0.680 & \textbf{0.735} & 0.700 & \textbf{0.756} \\
\textbf{Quality Score}  & 0.803 & \textbf{0.816} & 0.811 & \textbf{0.817} \\
\textbf{Final Score}    & 0.778 & \textbf{0.800} & 0.789 & \textbf{0.805} \\
\bottomrule
\end{tabular}}
\end{table}

\begin{table}[h]
\centering
\small
\caption{\textbf{Source of the Dynamic Degree trade-off.} We decompose Dynamic Degree across four settings on the full VBench evaluation set. On Wan2.1-1.3B, prompt rewriting alone accounts for most of the decrease, and full-video feedback regeneration, which uses the same VLM-based refinement step, exhibits a comparable drop. On CogVideoX-5B, our method preserves Dynamic Degree at the baseline level and outperforms regeneration.}
\label{tab:dynamic_degree}
\begin{tabular}{lcc}
\toprule
Method & CogVideoX-5B & Wan2.1-1.3B \\
\midrule
Original prompt (baseline)        & 0.542 & 0.542 \\
Prompt rewriting                  & 0.500 & 0.319 \\
Full-video feedback regeneration  & 0.458 & 0.431 \\
Original prompt + Ours            & 0.542 & 0.403 \\
\bottomrule
\end{tabular}
\end{table}

\begin{figure}[t]
\centering
\begin{tcolorbox}[
    colback=gray!5,
    colframe=gray!60,
    boxrule=0.5pt,
    arc=2pt,
    left=4pt,right=4pt,top=4pt,bottom=4pt,
    width=\linewidth
]
\begin{lstlisting}[style=promptstyle]
You are given sampled frames from an early-stage video generation preview.
This is not a final output.

Goal:
Improve overall consistency while preserving the original intent.

Inputs:
- original_prompt: "{prompt_en}"
- predicted_score_at_stop: {pred_score:.6f}
- score_range: 0.0 to 1.0 (higher is better)

Task:
- Evaluate the original prompt and preview frames together.
- Refine the prompt to better match the intended subject, action, and scene while fixing visible issues.

Guidelines:
- Base your revision on both the original prompt and what is visible in the preview.
- Make the prompt clearer, more concrete, and visually stable.
- Clarify subject visibility, action, framing, or scene when needed.
- Do not change the core meaning or add unrelated objects/actions.
- Keep the final prompt concise.

Negative prompt:
- Optional, single comma-separated string
- Short noun phrases only
- No instruction-style wording
- Maximum 5 items

Return ONLY valid JSON:
{
  "refined_prompt": "...",
  "negative_prompt": "..."
}
\end{lstlisting}
\end{tcolorbox}
\caption{\textbf{Prompt template for Visual Feedback Intervention (Trial~2).}
The VLM receives the original prompt, the decoded video preview, and the predicted preview score, and returns a refined prompt with an optional negative prompt in JSON format.}
\label{fig:prompt_template_refinement}
\end{figure}

\clearpage

\end{document}